\documentclass[10pt,twocolumn,letterpaper]{article}

\usepackage[pagenumbers]{cvpr} % To force page numbers, e.g. for an arXiv version

\definecolor{cvprblue}{rgb}{0.21,0.49,0.74}
\usepackage[pagebackref,breaklinks,colorlinks,allcolors=cvprblue]{hyperref}
\usepackage{booktabs}
\usepackage{multirow}
\usepackage{colortbl}
\usepackage[normalem]{ulem}

\usepackage[accsupp]{axessibility}

\useunder{\uline}{\ul}{}
\def\paperID{13209} % *** Enter the Paper ID here
\def\confName{CVPR}
\def\confYear{2025}

\title{Narrating the Video: Boosting Text-Video Retrieval via Comprehensive Utilization of Frame-Level Captions}

\makeatletter
\newcommand{\myfnsymbol}[1]{%
  \expandafter\@myfnsymbol\csname c@#1\endcsname
}
\newcommand{\@myfnsymbol}[1]{%
  \ifcase #1
  \or \TextOrMath{\textasteriskcentered}{*}% 1
  \or \TextOrMath{\textdagger}{\dagger}% 2
  \or \TextOrMath{\textdaggerdbl}{\ddagger}% 3
  \or \TextOrMath{\textbullet}{\bullet}% 4
  \fi
}
\newcommand{\equalcontributor}{\@myfnsymbol{1}}
\newcommand{\corresponding}{\@myfnsymbol{2}}
\newcommand{\workdone}{\@myfnsymbol{3}}
\makeatother

\author{%
Chan Hur\textsuperscript{\equalcontributor}
\and Jeong-hun Hong\textsuperscript{\equalcontributor} 
\and Dong-hun Lee 
\and Dabin Kang
\and Semin Myeong
\and Sang-hyo Park\textsuperscript{\corresponding}
\and Hyeyoung Park\textsuperscript{\corresponding} 
\and School of Computer Science and Engineering, Kyungpook National University \\
\begingroup
\hypersetup{urlcolor=magenta} % 이 부분에서 링크 색상을 분홍색으로 변경
\href{https://multimodal-understanding-group.github.io/NarVid/}{\tt\small https://multimodal-understanding-group.github.io/NarVid/}
\endgroup
}

\begin{document}
\maketitle
\renewcommand{\thefootnote}{\myfnsymbol{footnote}}

\footnotetext[1]{These authors contributed equally to this work.}%
\footnotetext[2]{Corresponding authors, equal leading contribution. }%
% \footnotetext[3]{Work done at Kyungpook National University.}

\setcounter{footnote}{0}% Restart footnote counter
\renewcommand{\thefootnote}{\fnsymbol{footnote}}
\begin{abstract}

In recent text-video retrieval, the use of additional captions from vision-language models has shown promising effects on the performance. However, existing models using additional captions often have struggled to capture the rich semantics, including temporal changes, inherent in the video. In addition, incorrect information caused by generative models can lead to inaccurate retrieval. To address these issues, we propose a new framework, Narrating the Video (NarVid), which strategically leverages the comprehensive information available from frame-level captions, the narration. The proposed NarVid exploits narration in multiple ways: 1) feature enhancement through cross-modal interactions between narration and video, 2) query-aware adaptive filtering to suppress irrelevant or incorrect information, 3) dual-modal matching score by adding query-video similarity and query-narration similarity, and 4) hard-negative loss to learn discriminative features from multiple perspectives using the two similarities from different views. Experimental results demonstrate that NarVid achieves state-of-the-art performance on various benchmark datasets. 
\end{abstract}
\section{Introduction}
\label{sec:intro}

\begin{figure}
    \centering
    \includegraphics[width=\linewidth]{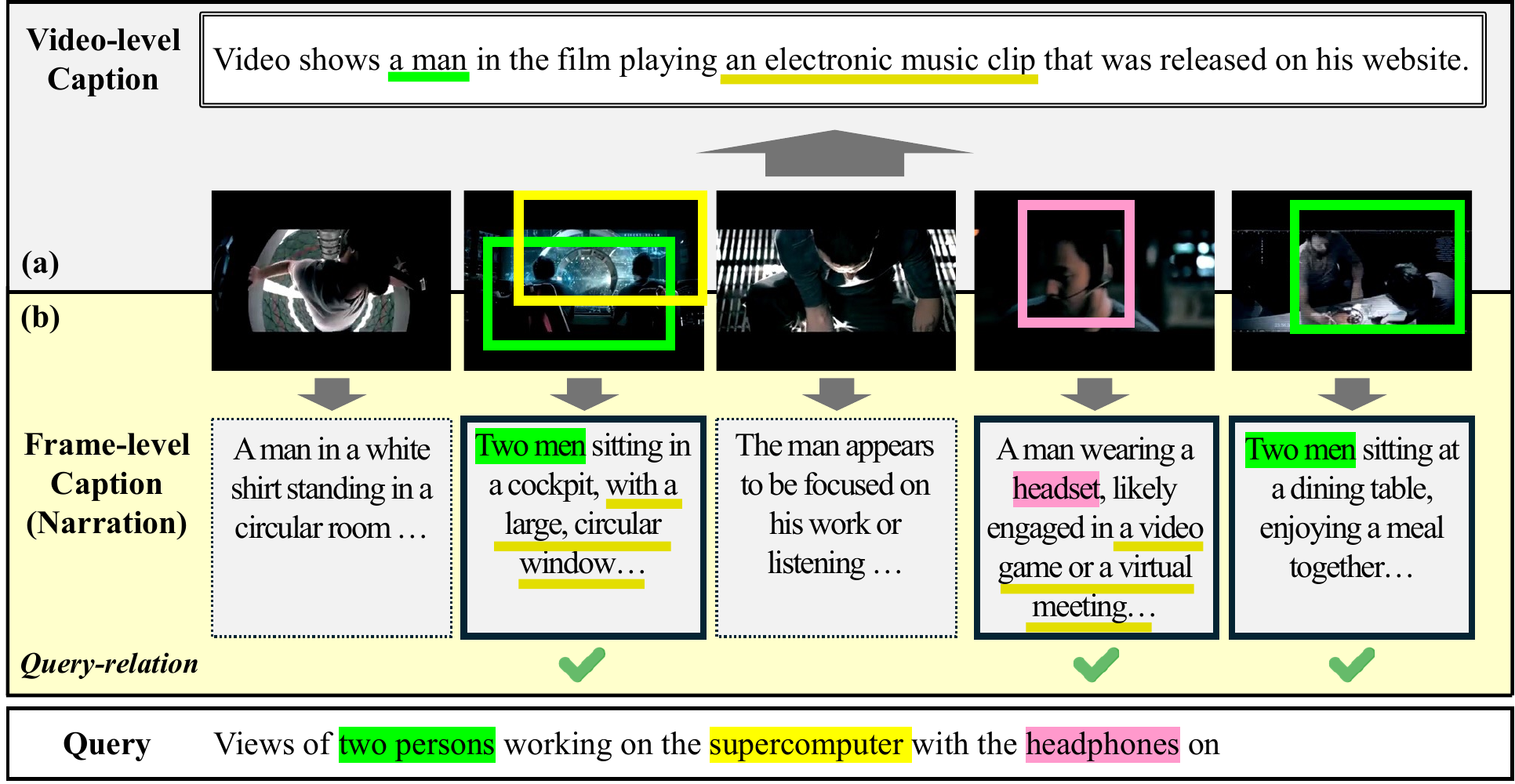}
    \vspace{-1.8em}
    \caption{(a) Unlike video-level caption information that summarizes the entire video, (b) the proposed framework utilizes frame-level captions to capture time-varying rich information and efficiently leverage selective information based on specific queries.}
    \vspace{-1em}
    \label{fig:fig1}
\end{figure}

With the increasing popularity of various video streaming platforms, text-to-video retrieval is emerging as an important topic in both academia and industry. In order to achieve accurate retrieval results, text-video retrieval encounters two challenging points: 1) a cross-modality gap from the heterogeneous data (vision and language) and 2) time-varying information in video. To address cross-modality challenges, many studies \cite{clip4clip, xclip, xpool, uatvr, tmass, frame-selection-study, rtq, stan, side4video, mpt, mmi-tvr} have leveraged vision-language pre-trained models such as CLIP \cite{clip} or BLIP \cite{blip} as a text and video backbone to resolve cross-modality gap between text-to-video domain. 
Moreover, to capture the time-varying relationships for video data, several studies \cite{clip4clip, stan, side4video, uatvr} have employed temporal modules, such as transformer encoder \cite{transformer} or 3D convolution \cite{3dconv}, over video frames.

Recently, some research \cite{cap4video,eavtr} has proposed to utilize generated captions from the video to bridge the gap between video and language. Cap4Video \cite{cap4video} derives video-level captions that summarize the entire video and then utilize the captions to mitigate the modality gap. However, as shown in \cref{fig:fig1}(a), a video-level caption may not reflect the contextual variations across the entire video. Furthermore, the usage of a single caption can lead to performance degradation when the generated caption is incorrectly obtained. EA-VTR \cite{eavtr} have proposed an event-aware method to capture time-varying events in the video by introducing frame-level captions, but they focus on training video encoder for event detection to address event-related video tasks, and does not consider query-based selective usage of the captions which is essential in the retrieval task.

To overcome these challenges in caption-based approaches, we propose a novel framework named Narrating the Video (NarVid), which generates frame-level captions across the video and utilizes them comprehensively throughout all stages of text-video retrieval. These frame-level captions obtained over time can be viewed as a story thus we call them a narration. As shown in \cref{fig:fig1}(b), instead of the existing video-level caption, frame-level captions allow us to capture diverse objects and attributes throughout the entire video. Meanwhile, since not all captions in the narration are relevant to the query, appropriate handling of the generated information is required.

In our NarVid framework, we utilize the narration in a variety of ways to improve retrieval performance. First, we extract mutually enhanced features through co-attention between the pairs of video frames and the generated captions. The enhanced frame-level features are further refined by filtering out irrelevant frames and incorrectly generated captions based on similarities to the given query. The narration features are also used in the matching process, where query-video similarity and query-narration similarity are fused to provide better retrieval results. The query-video-narration relationship is also applied to define a novel hard negative loss for learning more discriminative features. The proposed comprehensive utilization of narration throughout the retrieval process successfully reduces the cross-modality gap, leading to state-of-the-art results in text-video retrieval.

To summarize, our contributions are as follows:
\begin{enumerate}
    \item We propose a novel text-video retrieval framework, NarVid, which strategically leverages frame-level captions obtained from the generative model. These frame-level captions, called narration, could provide time-varying rich information to boost the retrieval performance.

    \item Our NarVid framework comprehensively utilizes narration in the entire matching process through four modules: 1) enhancing cross-modal features, 2) filtering irrelevant or incorrect features, 3) complementing matching scores, and 4) learning with hard-negative sampling.

    \item We validate the performance of NarVid through extensive experiments on four benchmark datasets. The proposed model outperforms state-of-the-art models in retrieval performances on MSR-VTT (52.7\%), MSVD (53.1\%), VATEX (68.4\%), and DiDeMo (53.4\%).
\end{enumerate}

%-------------------------------------------------------------------------
\section{Related works}
\label{sec:related}

%-------------------------------------------------------------------------
\subsection{Text-Video Retrieval}

Vision-language pre-trained models such as CLIP \cite{clip} have demonstrated the feasibility of learning a multimodal embedding space through contrastive learning on large-scale data, exhibiting strong generalization performance. Building on this success, recent research has focused on extending CLIP to video domain tasks, such as text-video retrieval. CLIP4Clip \cite{clip4clip} successfully adapted CLIP for text-video retrieval by extracting features from individual video frames and aggregating them. Following CLIP4Clip, X-CLIP \cite{xclip} enhanced the features of each modality by integrating both coarse-grained and fine-grained features. Additionally, X-Pool \cite{xpool} utilized an attention mechanism to highlight frames that hold the most relevance to the text. Recently, probabilistic approaches have emerged to capture variability in the data. UAVTR \cite{uatvr} introduced a technique to represent the pairs of text-video features as probability distributions. T-MASS \cite{tmass} presents an approach by modeling text as a stochastic embedding, enriching the semantic representation. Although these CLIP-based retrieval models have shown significant performance improvements, they have limitations in bridging the substantial domain gap between video and text. Our paper mitigates this limitation by employing auxiliary information from the generative model and utilizing the generated information in diverse ways.

%-------------------------------------------------------------------------
\subsection{Vision-Language Model with Generated Information}

Recent work on vision-language models has shown that captions generated using large language models can improve feature representation. In the image domain, ALIP \cite{alip} introduced triplet similarities among images, text, and generated captions to improve text-image retrieval performance. LaCLIP \cite{laclip} used rewritten textual descriptions to augment CLIP learning for obtaining enhanced multimodal features. In the video domain, recent studies have enhanced cross-modal interactions by utilizing additional data, such as captions derived from videos. CoVR \cite{covr} improved the composed video retrieval performance by using cross-attention between videos and generated captions, and Cap4Video \cite{cap4video} proposed a co-attention framework using single-sentence captions summarizing entire videos.
Nonetheless, these studies primarily focus on global video content, often overlooking local details. In this context, EA-VTR \cite{eavtr} uses frame-level captions in defining the loss function of the encoder by comparing every possible pair of frames and captions from all videos, which may cause undesirable effects by unrefined information. Our approach strategically utilizes frame-level captions, including query-aware filtering methods, to emphasize the temporal aspects of the videos and capture local details, reducing the domain gap between video and text.

%-------------------------------------------------------------------------
\begin{figure*}
    \centering
    \includegraphics[width=\linewidth]{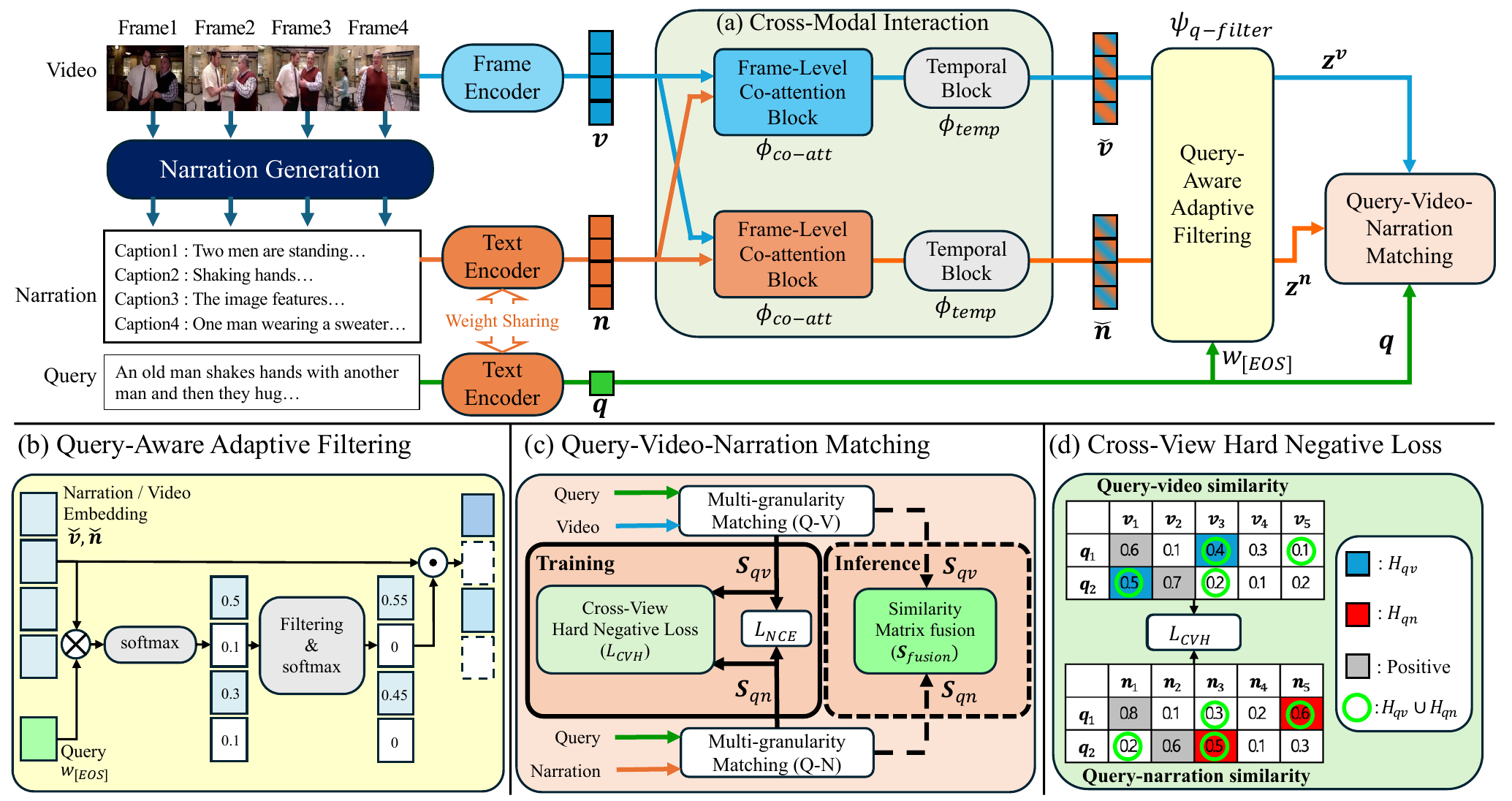}
    \vspace{-2em}
    \caption{An overview of the proposed framework, NarVid. The method first generates frame-level captions (narration) for each video. (a) Using the frame-level features of the video and narration, enhanced features are obtained through cross-modal interaction with co-attention and temporal block. (b) These enhanced features are further refined using query-aware adaptive filtering. (c) Then, the query-video and query-narration similarity matrices obtained through the multi-granularity matching are utilized for training and inference. (d) To enhance the discriminative ability of the model, we additionally use a cross-view hard negative loss during training.}
    \vspace{-1em}
    \label{fig:framework}
\end{figure*}

\section{Proposed Methods}
\label{sec:method}

%-------------------------------------------------------------------------
\subsection{Overall Process}

\textbf{Problem formulation.} Given a query text, the text-to-video retrieval task is to identify the most semantically relevant video from a set of candidate videos \( V \). The task is formally defined as the optimization problem of finding the optimal video \( \boldsymbol{v^*} \) that maximizes the similarity with the given query \( \boldsymbol{q} \), which can be written as
\begin{equation}
  \boldsymbol{v^*} = \arg\max_{\boldsymbol{v} \in V} s(\boldsymbol{q}, \boldsymbol{v}).
  \label{eq:preliminary}
\end{equation}
Here, \( s(\boldsymbol{q}, \boldsymbol{v}) \) is the similarity function that needs to be modeled appropriately to achieve good retrieval performance.

\textbf{Pipeline.} The proposed framework for calculating similarity follows the process outlined in \cref{fig:framework}. A query and a video are represented as \( \boldsymbol{q} = (w_1, \dots, w_L, w_{[EOS]}) \) and \( \boldsymbol{v} = (f_1, \dots, f_K) \) using clip encoders, where \( w_L \) is the last word of the query, \( w_{[EOS]} \) denotes a special token representing the query's global meaning, and \( f_K \) is the last frame of the video. Additionally, we generate a sequence of captions for individual frames using the narration generation module, represented as \( \boldsymbol{n} = (c_1, \dots, c_K) \) (\cref{sec:method:narration-generation}). The encoded features of the narration and video are enhanced through mutual co-attention and temporal transformation (\cref{sec:method:cross-modal-inter}). Query-aware adaptive filtering refines these features based on the query (\cref{sec:method:filtering}), then query-video and query-narration matching is employed to obtain the final similarity score (\cref{sec:method:matching}). A cross-view hard negative loss is also defined by combining query-video and query-narration similarities to enhance retrieval performance (\cref{sec:method:train-infer:objective}).

%-------------------------------------------------------------------------
\subsection{Narration Generation from Video}
\label{sec:method:narration-generation}

In this paper, we try to solve some limitations of the previous work \cite{cap4video} that uses a video-level caption to describe a video. Since a video-level caption is usually generated by focusing on the key scenes, it is likely to miss other narrative flows in the video clip. In addition, when a single generated caption is used to represent the entire video frames, any errors (\eg, hallucinations) in the generation model can severely affect the retrieval performance.
To address these issues, we propose the concept of narration. For each image frame \( f_k \) in a video \( \boldsymbol{v} = (f_1, \dots, f_K) \), we apply a large multimodal model (LMM) \cite{llava} to independently interpret it and generate a caption \( c_k \), so that we have a sequence of captions, \( \boldsymbol{n} = (c_1, \dots, c_K) \). Note that our framework does not focus on any specific caption generation model, the choice of captioning model and prompts can be flexible, considering various environments. The generated captions arranged in chronological order can linguistically express the semantic changes in the video over time. The generated narration is subsequently used in various ways in our proposed NarVid to improve the performance of text-to-video retrieval.

%-------------------------------------------------------------------------
\subsection{Video-Narration Cross-Modal Interaction}
\label{sec:method:cross-modal-inter}

Each caption \( c_k \) in the generated narration can be paired with a video frame \( f_k \) to interact and enhance mutual information. Specifically, we use the co-attention structure \cite{vilbert} for interactions, a method commonly employed for cross-modal information enrichment. \cref{fig:framework} (a) shows the cross-modal interaction between each video \( \boldsymbol{v} \) and its corresponding narration \( \boldsymbol{n} \). In this context, the transformer blocks perform co-attention operations with cross-modal inputs to obtain the enhanced features \( \boldsymbol{\hat{v}} \) and \( \boldsymbol{\hat{n}} \), defined as
\begin{equation}
  (\boldsymbol{\hat{v}}, \boldsymbol{\hat{n}}) = \phi_{co-att}(\boldsymbol{v}, \boldsymbol{n}).
  \label{eq:co-attention}
\end{equation}
Through co-attention between video frames and their corresponding captions instead of a video-level caption, we can naturally capture intricate relationships between specific frames and their narrative context. This approach enhances features by leveraging rich information from all frames through individual frame-caption pairs. 

The obtained sequential features \( \boldsymbol{\hat{v}} = (\hat{f}_1, \dots, \hat{f}_K) \) and \( \boldsymbol{\hat{n}} = (\hat{c}_1, \dots, \hat{c}_K) \) are then passed through a temporal block \cite{clip4clip} to capture temporal relationships in the sequence of each modality, producing temporally enhanced features as follows:
\begin{equation}
  \boldsymbol{\check{v}} = \phi_{temp}(\boldsymbol{\hat{v}}), \quad \boldsymbol{\check{n}} = \phi_{temp}(\boldsymbol{\hat{n}}).
  \label{eq:temporal}
\end{equation}
Since the enhanced features \( \boldsymbol{\check{v}} \) and \( \boldsymbol{\check{n}} \) are still in sequential forms with frame-level features, we can apply a filtering process to select query-relevant features.

%-------------------------------------------------------------------------
\subsection{Query-Aware Adaptive Filtering}
\label{sec:method:filtering}

To filter out query-irrelevant frames or erroneous captions in the narration, we propose a query-aware filtering module. Conventional video frame filtering methods employ a top-k approach \cite{frame-selection-study, adaclip}, selecting the top-k frames with the highest similarity to the query. However, using a fixed  for filtering may cause a loss of important frames or the inclusion of irrelevant ones. To overcome this limitation, we adopt the nucleus sampling method \cite{nucleus} to implement nucleus filtering, allowing the adaptive determination of the number of keyframes.

\cref{fig:framework} (b) illustrates our nucleus filtering process. For a given query \( \boldsymbol{q} = (w_1, \dots, w_L, w_{[EOS]}) \) and the enhanced features \( \boldsymbol{\check{v}} = (\check{f}_1, \dots, \check{f}_K) \) and \( \boldsymbol{\check{n}} = (\check{c}_1, \dots, \check{c}_K) \), we first compute the frame-level similarities between \( w_{[EOS]} \) and the features \( \check{f}_k \) and \( \check{c}_k \) (\( k = 1, \dots, K \)). We then apply the softmax function to normalize these similarity values to the range \([0, 1]\). In the feature selection phase, we start with the feature having the highest similarity and add features one by one until the cumulative similarity exceeds a predefined threshold \( p \). Through this process, we obtain a pair of final sequences of features, \( \boldsymbol{z}^v \) for video and \( \boldsymbol{z}^n \) for narration, defined as
\begin{equation}
  (\boldsymbol{z}^v, \boldsymbol{z}^n) = \psi_{q-filter}(\boldsymbol{\check{v}}, \boldsymbol{\check{n}}; \boldsymbol{q}).
  \label{eq:4}
\end{equation}
This adaptive filtering process effectively selects key features for video and narration that are highly relevant to the specific query.

%-------------------------------------------------------------------------
\subsection{Query-Video-Narration Matching}
\label{sec:method:matching}
The filtered feature vectors are used in a detailed multi-granularity matching process using coarse and fine granularities \cite{ucofia, pidro}. In addition to traditional inter-modality matching between query and video, we propose an additional intra-modality matching between query and narration to improve retrieval performance.

Coarse-grained matching is a common method for matching overall semantics between query and video. In addition to query-video matching, a similar query-narration matching process is conducted in NarVid. Specifically, we first apply weighted pooling to the sequence of features for the video \( \boldsymbol{z}^v \) and narration \( \boldsymbol{z}^n \) to obtain their coarse representations, where the weights are determined by the similarity scores from query-aware filtering module (\cref{sec:method:filtering}). The \( w_{[EOS]} \) token is used for the coarse representation of the query \( \boldsymbol{q} \). Then, the coarse-grained matching scores \( s_{coarse}(\boldsymbol{q}, \boldsymbol{z}^v) \) and \( s_{coarse}(\boldsymbol{q}, \boldsymbol{z}^n) \) are calculated using cosine similarity between their coarse representations, effectively leveraging the filtering scores as weights in the pooling process.

Fine-grained matching is a detailed matching strategy for capturing fine-level relationships between query and video. We extend the conventional method to include query-narration matching. Specifically, the query-video matching score \( s_{fine}(\boldsymbol{q}, \boldsymbol{z}^v) \) is defined as the maximum cosine similarity between frame and word embeddings of the query, following the approach in Cap4Video \cite{cap4video}. In NarVid, we also define the query-narration matching score \( s_{fine}(\boldsymbol{q}, \boldsymbol{z}^n) \) as the maximum similarity between the embeddings of frame-level captions and words.

By averaging the coarse and fine scores, we obtain two multi-grained similarity scores, \( s_{qv} \) for query-video matching and \( s_{qn} \) for query-narration matching, which are used for training and inference. For more detailed information is qiven in Supplementary \cref{sec:supple:multi-granularity} % Sec. A

%-------------------------------------------------------------------------
\subsection{Training and Inference}
\label{sec:method:train-infer}
\subsubsection{Training Objective}
\label{sec:method:train-infer:objective}
\textbf{Contrastive loss.} For a given batch set \( \{(\boldsymbol{q}_i, \boldsymbol{v}_i, \boldsymbol{n}_i)\}_{i=1}^B \), we calculate the query-video similarity matrix \( \boldsymbol{S}_{qv} \) and the query-narration similarity matrix \( \boldsymbol{S}_{qn} \), where each element is defined as
\(
\boldsymbol{S}_{qv}(i, j) = s_{qv}(\boldsymbol{q}_i, \boldsymbol{v}_j), \quad \boldsymbol{S}_{qn}(i, j) = s_{qn}(\boldsymbol{q}_i, \boldsymbol{n}_j).
\)
We utilize these similarity matrices to enhance retrieval performance by considering both inter-modality (query-video) and intra-modality (query-narration) aspects. This approach modifies the InfoNCE loss \cite{infonce} commonly used in text-video retrieval to magnify the discrepancy between positive and negative samples from both perspectives. The modified loss functions are defined as
\begin{align} 
    &L_{qv}=-\frac{1}{2B} \sum_{i=1}^B \left\{ \log \frac{e^{\boldsymbol{S}_{qv}(i, i)}}{\displaystyle\sum_{j=1}^B e^{\boldsymbol{S}_{qv}(i, j)}} + \log \frac{e^{\boldsymbol{S}_{qv}(i, i)}}{\displaystyle\sum_{j=1}^B e^{\boldsymbol{S}_{qv}(j, i)}} \right\}, \nonumber
\\
    &L_{qn}=-\frac{1}{2B} \sum_{i=1}^B \left\{ \log \frac{e^{\boldsymbol{S}_{qn}(i, i)}}{\displaystyle\sum_{j=1}^B e^{\boldsymbol{S}_{qn}(i, j)}} + \log \frac{e^{\boldsymbol{S}_{qn}(i, i)}}{\displaystyle\sum_{j=1}^B e^{\boldsymbol{S}_{qn}(j, i)}} \right\}, \nonumber
\\
        &L_{NCE} = \frac{1}{2} (L_{qv} + L_{qn}).
    \label{eq:infonce}
\end{align}

\textbf{Cross-view hard negative loss.} Hard negative samples in contrastive learning mean negative samples that closely resemble positive samples, challenging the model to distinguish between them. Although several studies \cite{cross_encoder_good_teacher, contrastive_hard_negative} showed that more discriminative features can be obtained if such hard negative samples are well utilized for training, they exclusively focused on either inter-modality or intra-modality when selecting hard negatives. In this paper, we propose a novel hard negative loss that leverages both different views: inter-modality and intra-modality. 
As shown in \cref{fig:framework} (d)
using the similarity matrices \( \boldsymbol{S}_{qv} \) and \( \boldsymbol{S}_{qn} \), we define two index sets of hard negatives for a given query \( \boldsymbol{q}_i \) from the views of inter-modality and intra-modality. these sets are defined as
\begin{align}
  H_{qv}(i) &= \{j \mid \boldsymbol{S}_{qv}(i, i) - \boldsymbol{S}_{qv}(i, j) < \lambda \sigma_i^{qv}, j \neq i \}, \nonumber \\
  H_{qn}(i) &= \{j \mid \boldsymbol{S}_{qn}(i, i) - \boldsymbol{S}_{qn}(i, j) < \lambda \sigma_i^{qn}, j \neq i \}.
  \label{eq:hard_selection}
\end{align}
Here, \( \sigma_i^{qv} \) and \( \sigma_i^{qn} \) are the standard deviation of the values in \( i \)-th row of \( \boldsymbol{S}_{qv} \) and \( \boldsymbol{S}_{qn} \), and \( \lambda \) is a hyperparameter. The unified hard negative index for \( i \)-th query, \( H_i \), is obtained by combining both sets:
\begin{equation}
  H_i = H_{qv}(i) \cup H_{qn}(i).
  \label{eq:6}
\end{equation}
We can apply the same process for the transposed matrices, \( \boldsymbol{S}_{qv}^T \) and \( \boldsymbol{S}_{qn}^T \) to obtain hard negatives for \( i \)-th video \( \boldsymbol{v}_i \) and narration \( \boldsymbol{n}_i \), denoting these as \( H_i^T \).
Subsequently, we implement a hard negative rank loss, built on the concept of hinge loss \cite{rankloss}, using the index sets of unified hard negatives. Specifically, using the threshold value \( \lambda \sigma_i^{qv} \) scaled with \( \eta \), the hard negative loss for query-video pairs is defined as
\begin{align}
  L_{qv}^{hard} = \frac{1}{2B} \sum_{i=1}^B \left( \sum_{j \in H_i} \Delta_{qv}(i, j) + \sum_{j \in H_i^T} \Delta_{vq}(i, j) \right), \nonumber
\end{align}
where
\begin{align}
    \Delta_{qv}(i, j) &= \max(0, \boldsymbol{S}_{qv}(i, j) - \boldsymbol{S}_{qv}(i, i) + \eta \lambda \sigma_i^{qv}), \nonumber \\
    \Delta_{vq}(i, j) &= \max(0, \boldsymbol{S}_{qv}(j, i) - \boldsymbol{S}_{qv}(i, i) + \eta \lambda \sigma_i^{vq}).
\label{eq:hard_rank_loss}
\end{align}

Similarly, We can define the hard negative loss \( L_{qn}^{hard} \) for query-narration pairs using the similarity matrices \( \boldsymbol{S}_{qn} \). The overall cross-view hard negative loss is then defined as
\begin{align}
  L_{CVH} = L_{qv}^{hard} + L_{qn}^{hard},
  \label{eq:9}
\end{align}
and the final training loss is defined as 
\begin{align}
  L = L_{NCE} + \alpha L_{CVH},
  \label{eq:10}
\end{align}
where \( \alpha \) is a hyperparameter.

\subsubsection{Inference Pipeline}
\label{sec:method:train-infer:pipeline}
Since the elements in the two similarity matrices \( \boldsymbol{S}_{qv} \) and \( \boldsymbol{S}_{qn} \) often exist in different ranges of values, the direct summation of the two matrices can disproportionately emphasize the narration aspect. To mitigate this problem, the element values in each matrix are separately standardized before summation, using their means \( \mu^{qv} \) and \( \mu^{qn} \) and the standard deviations \( \sigma^{qv} \) and \( \sigma^{qn} \). For inference, the two standardized matrices are then fused to form the final score matrix \( \boldsymbol{S}_{fusion} \), ensuring balanced consideration of both aspects, which can be defined as
\begin{align}
  \boldsymbol{S}_{fusion}(i, j) = \frac{\boldsymbol{S}_{qv}(i, j) - \mu^{qv}}{\sigma^{qv}} + \frac{\boldsymbol{S}_{qn}(i, j) - \mu^{qn}}{\sigma^{qn}}.
  \label{eq:infer}
\end{align}
\begin{table*}[]
\centering
\resizebox{0.9\textwidth}{!}{%
\begin{tabular}{@{}llcccccccccc@{}}
\toprule[1.5pt]
                         &                                 & \multicolumn{5}{c}{\textbf{Text-to-Video}}                                                        & \multicolumn{5}{c}{\textbf{Video-to-Text}}                                  \\ \cmidrule(l){3-12} 
\multirow{-2}{*}{\textbf{Method}} & \multirow{-2}{*}{\textbf{Venue}}         & R@1           & R@5           & R@10          & MdR          & MnR                                & R@1           & R@5           & R@10          & MdR          & MnR          \\ \midrule
\multicolumn{12}{l}{\cellcolor[HTML]{D9D9D9}\textit{CLIP-ViT-B/32}}                                                                                                                                                                          \\
CLIP4Clip \cite{clip4clip}                & \multicolumn{1}{l|}{Neurocomputing'22}   & 44.5          & 71.4          & 81.6          & 2.0          & \multicolumn{1}{c|}{15.3}          & 42.7          & 70.9          & 80.6          & 2.0          & 11.6         \\
X-CLIP \cite{xclip}                  & \multicolumn{1}{l|}{MM'22}      & 46.1          & 73.0          & 83.1          & 2.0          & \multicolumn{1}{c|}{13.2}          & 46.8          & 73.3          & 84.0          & 2.0          & 9.1          \\
X-Pool \cite{xpool}                   & \multicolumn{1}{l|}{CVPR'22}    & 46.9          & 72.8          & 82.2          & 2.0          & \multicolumn{1}{c|}{14.3}          & 44.4          & 73.3          & 84.0          & 2.0          & 9.0          \\
UATVR \cite{uatvr}                   & \multicolumn{1}{l|}{ICCV'23}    & 47.5          & 73.9          & 83.5          & 2.0          & \multicolumn{1}{c|}{12.3}          & 46.9          & 73.8          & 83.8          & 2.0          & 8.6          \\
TEFAL \cite{tefal}                   & \multicolumn{1}{l|}{ICCV'23}    & 49.4          & 75.9          & 83.9          & 2.0          & \multicolumn{1}{c|}{-}             & -             & -             & -             & -            & -            \\
PAU \cite{pau}                     & \multicolumn{1}{l|}{NeurIPS'23} & 48.5          & 72.7          & 82.5          & 2.0          & \multicolumn{1}{c|}{14.0}          & -             & -             & -             & -            & -            \\
Cap4Video \cite{cap4video}                & \multicolumn{1}{l|}{CVPR'23}    & 49.3          & 74.3          & 83.8          & 2.0          & \multicolumn{1}{c|}{12.0}          & 47.1          & 73.7          & 84.3          & 2.0          & 8.7          \\
MPT \cite{mpt}                    & \multicolumn{1}{l|}{MM'24}   & 46.3          & 70.9          & 80.7          & -          & \multicolumn{1}{c|}{15.6} & 45.0          & 70.9          & 80.6          & -          & 12.7          \\
ProTA \cite{prota}                   & \multicolumn{1}{l|}{ICME'24}   & 48.1          & 75.4          & 84.3          & 2.0          & \multicolumn{1}{c|}{12.5}          & 45.9          & 75.5          & 84.6          & 2.0          & 9.0          \\
T-MASS \cite{tmass}                  & \multicolumn{1}{l|}{CVPR'24}    & 50.2          & 75.3          & 85.1          & \textbf{1.0}          & \multicolumn{1}{c|}{11.9}          & 47.7          & \textbf{78.0} & \textbf{86.3} & 2.0          & 8.0          \\ \midrule
NarVid                   & \multicolumn{1}{l|}{}           & \textbf{51.0} & \textbf{76.4} & \textbf{85.2} & \textbf{1.0} & \multicolumn{1}{c|}{\textbf{11.6}} & \textbf{50.0} & 75.4          & 83.8          & \textbf{1.5} & \textbf{7.9} \\ \midrule
\multicolumn{12}{l}{\cellcolor[HTML]{D9D9D9}\textit{CLIP-ViT-B/16}}                                                                                                                                                                          \\
X-CLIP \cite{xclip}                  & \multicolumn{1}{l|}{MM'22}      & 49.3          & 75.8          & 84.8          & 2.0          & \multicolumn{1}{c|}{12.2}          & 48.9          & 76.8          & 84.5          & 2.0          & 8.1          \\
UATVR \cite{uatvr}                   & \multicolumn{1}{l|}{ICCV'23}    & 50.8          & 76.3          & 85.5          & \textbf{1.0}          & \multicolumn{1}{c|}{12.4}          & 48.1          & 76.3          & 85.4          & 2.0          & 8.0          \\
TEFAL \cite{tefal}                   & \multicolumn{1}{l|}{ICCV'23}    & 49.9          & 76.2          & 84.4          & 2.0          & \multicolumn{1}{c|}{11.4}          & -             & -             & -             & -            & -            \\
Cap4Video \cite{cap4video}               & \multicolumn{1}{l|}{CVPR'23}    & 51.4          & 75.7          & 83.9          & \textbf{1.0}          & \multicolumn{1}{c|}{12.4}          & 49.0          & 75.2          & 85.0          & 2.0          & 8.0          \\
MPT \cite{mpt}                    & \multicolumn{1}{l|}{MM'24}   & 49.2          & 72.9          & 82.4          & -          & \multicolumn{1}{c|}{15.5} & 47.4          & 73.9          & 83.4          & -          & 10.9          \\
ProTA \cite{prota}                    & \multicolumn{1}{l|}{ICME'24}   & 50.9          & 77.0          & 85.4          & \textbf{1.0}          & \multicolumn{1}{c|}{11.1} & 48.5          & 77.0          & 87.0          & 2.0          & 7.9          \\
T-MASS \cite{tmass}                   & \multicolumn{1}{l|}{CVPR'24}    & \textbf{52.7}          & 77.1          & \textbf{85.6}          & \textbf{1.0}          & \multicolumn{1}{c|}{\textbf{10.5}}          & 50.9          & \textbf{80.2} & \textbf{88.0} & \textbf{1.0}          & \textbf{7.4} \\ \midrule
NarVid                   & \multicolumn{1}{l|}{}           & \textbf{52.7} & \textbf{77.7} & \textbf{85.6} & \textbf{1.0} & \multicolumn{1}{c|}{12.3}          & \textbf{51.1} & 76.8          & 85.2          & \textbf{1.0} & 9.0          \\ \bottomrule[1.5pt]
\end{tabular}
}
\vspace{-0.5em}
\caption{Retrieval results comparisons on MSR-VTT. Bold denotes the best performance. Results are reported without any post-processing such as QB-Norm \cite{qbnorm} or DSL \cite{dsl}.}
\vspace{-0.5em}
\label{tab:tab1}
\end{table*}

\begin{table*}[t]
\centering
\resizebox{0.9\textwidth}{!}{%
\begin{tabular}{@{}llccccccccc@{}}
\toprule[1.5pt]
\multirow{2}{*}{\textbf{Method}}    & \multirow{2}{*}{\textbf{Venue}}          & \multicolumn{3}{c}{\textbf{MSVD}}                                  & \multicolumn{3}{c}{\textbf{VATEX}}                                 & \multicolumn{3}{c}{\textbf{DiDeMo}}           \\ \cmidrule(l){3-11} 
                           &                                 & R@1           & R@5           & R@10                               & R@1           & R@5           & R@10                               & R@1           & R@5           & R@10          \\ \midrule
CLIP4Clip \cite{clip4clip} & \multicolumn{1}{l|}{Neurocomputing'22}   & 45.2          & 75.5          & \multicolumn{1}{c|}{84.3}          & 55.9          & 89.2          & \multicolumn{1}{c|}{95.0}          & 42.8          & 68.5          & 79.2          \\
UATVR \cite{uatvr}         & \multicolumn{1}{l|}{ICCV'23}    & 49.7          & 79.0          & \multicolumn{1}{c|}{87.3}          & 64.5          & 92.6          & \multicolumn{1}{c|}{96.8}          & 45.8          & 73.7          & 83.3          \\
TEFAL \cite{tefal}         & \multicolumn{1}{l|}{ICCV'23}    & -             & -             & \multicolumn{1}{c|}{-}             & 61.0          & 90.4          & \multicolumn{1}{c|}{95.3}          & -             & -             & -             \\
PAU \cite{pau}             & \multicolumn{1}{l|}{NeurIPS'23} & 47.3          & 77.4          & \multicolumn{1}{c|}{85.5}          & -             & -             & \multicolumn{1}{c|}{-}             & 48.6          & 76.0          & 84.5          \\
Cap4Video \cite{cap4video} & \multicolumn{1}{l|}{CVPR'23}    & 51.8          & 80.8          & \multicolumn{1}{c|}{88.3}          & 66.6          & 93.1          & \multicolumn{1}{c|}{97.0}          & 52.0          & 79.4          & 87.5          \\
MPT \cite{mpt}         & \multicolumn{1}{l|}{MM'24}   & -             & -             & \multicolumn{1}{c|}{-}             & -             & -             & \multicolumn{1}{c|}{-}             & 46.4          & 72.2          & 81.4          \\
ProTA \cite{prota}         & \multicolumn{1}{l|}{ICME'24}   & -             & -             & \multicolumn{1}{c|}{-}             & -             & -             & \multicolumn{1}{c|}{-}             & 47.2          & 74.6          & 83.0          \\
T-MASS \cite{tmass}        & \multicolumn{1}{l|}{CVPR'24}    & -             & -             & \multicolumn{1}{c|}{-}             & 65.6          & 93.9          & \multicolumn{1}{c|}{\textbf{97.2}} & 53.3          & \textbf{80.1} & \textbf{87.7} \\ \midrule
NarVid                     & \multicolumn{1}{l|}{}           & \textbf{53.1} & \textbf{81.4} & \multicolumn{1}{c|}{\textbf{88.8}} & \textbf{68.4} & \textbf{94.0} & \multicolumn{1}{c|}{97.1}          & \textbf{53.4} & 79.1          & 86.3          \\ \bottomrule[1.5pt]
\end{tabular}
}
\vspace{-0.5em}
\caption{Text-to-video retrieval results on MSVD, VATEX, and DiDeMo. Bold denotes the best performance. Results are reported without any post-processing such as QB-Norm \cite{qbnorm} or DSL \cite{dsl}.}
\vspace{-1em}
\label{tab:tab2}
\end{table*}
\section{Experimental Results}
\label{sec:experiments}

%-------------------------------------------------------------------------
\subsection{Experimental Settings}

\textbf{Datasets and metrics.} We validate our framework on four benchmark datasets using the same protocol as in Cap4Video \cite{cap4video}. MSR-VTT \cite{msrvtt} consists of 10K videos, each containing approximately 20 captions, and we utilize the 1k-A test split \cite{jsfusion} for evaluation. MSVD \cite{msvd} includes 1,970 videos, with each video accompanied by an average of 40 captions. VATEX \cite{vatex} comprises 35K video clips, each featuring multilingual captions. DiDeMo \cite{didemo} consists of 10K videos annotated with around 40K captions, where we evaluate video-paragraph retrieval by concatenating all sentence descriptions for each video into a single query. 

For evaluation, we employ standard retrieval metrics: Recall at Rank K (R@K), Median Rank (MdR), and Mean Rank (MnR). Higher values for R@K indicate better performance, while lower values for both MdR and MnR are preferred.

\textbf{Implementation details.} We use the baseline settings of CLIP4Clip \cite{clip4clip} with ViT-B/16 \cite{vit} as the visual backbone and the CLIP text encoder for query and narration encoding. For a detailed analysis of MSR-VTT, we additionally use ViT-B/32 as a visual encoder. To obtain the narration, LLaVa 1.5 7B \cite{llava} is used as the baseline generative model. We use a batch size of 64 and train for 5 epochs on three datasets, except for DiDeMo, where a batch size of 24 and 10 learning epochs are used. The threshold \( p \) used in query-aware filtering is set to 0.4 for ViT-B/32 and 0.5 for ViT-B/16. All experiments are conducted on two NVIDIA RTX A6000 GPUs. Additional implementation details can be found in Supplementary \cref{sec:supple:experimental-settings:implementation_details} %Sec. B.2

%-------------------------------------------------------------------------
\subsection{Comparison with State-of-the-Arts}

We compare NarVid with state-of-the-art models across the four benchmark datasets. 

\textbf{MSR-VTT dataset.}
\cref{tab:tab1} shows the superior retrieval performance of NarVid compared to other methods across the overall recall metrics. Compared with the baseline CLIP4Clip, without using any generation information, NarVid achieves a dramatic improvement across all retrieval metrics. The result shows the impressive effect of the utilization of narration. Furthermore, NarVid also outperforms Cap4Video, which uses video-level captions, on both ViT-B/32 and ViT-B/16 backbones across most evaluation metrics, including video-to-text retrieval. These results highlight the benefits of the proposed comprehensive utilization of frame-level captions.
We also compare NarVid to the recent state-of-the-art model T-MASS \cite{tmass}, which uses stochastic text embedding as a training strategy. Although NarVid outperforms T-MASS on R@1 across all datasets, it shows decreased performances on R@5 and R@10. We conclude that these performance gaps can be attributed to the stochastic embedding methods proposed by T-MASS, which augment the number of matched text-video pairs. Such stochastic embedding methods in the matching phase could be partially integrated into NarVid as an additional training strategy to improve performance in future work.

\cref{tab:tab2} presents the comparative results for the MSVD, VATEX, and DiDeMo datasets. 

\textbf{MSVD dataset.} We demonstrate clear performance improvements across all recall metrics. However, when compared to other datasets, the performance improvement in R@1 over the baseline model (CLIP4Clip) is relatively modest (MSVD +7.9\%, VATEX +12.5\%, and DiDeMo +10.2\%). It may be attributed to the simplicity and brevity of MSVD queries. We also observed that some queries could be matched to multiple videos in a test set, which may limit the effectiveness of narration information. This tendency is observed in the results of the MSR-VTT dataset with similar forms of queries.

\textbf{VATEX dataset.} NarVid exhibits a significant improvement in R@1 compared to the baseline. Due to the properties of the VATEX dataset, the query sentences are relatively long and contain words describing various attributes. Such sentences can be well matched with the diverse information included in the narration, which can lead to a substantial improvement in retrieval performance.

\textbf{DiDeMo dataset.} NarVid demonstrates superior performance in R@1 compared to other models; however, the performance gap is smaller than observed in other datasets. This reduced margin is likely due to the static nature of DiDeMo videos, which diminishes the effectiveness of the rich temporal information that narration provides.

In summary, our method achieves state-of-the-art results across all four benchmarks, demonstrating the effectiveness of the narration-based approach in text-video retrieval.

%-------------------------------------------------------------------------

\begin{table}
\centering
\resizebox{0.8\columnwidth}{!}{%
\begin{tabular}{@{}cccc|ccc@{}}
\toprule[1.5pt]
\multicolumn{4}{c|}{Proposed   modules} & \multicolumn{3}{c}{Text-to-Video}    \\ \midrule
NM       & CMI       & QAF      & CVH      & R@1           & R@5           & R@10          \\ \midrule
         &          &         &         & 44.5          & 71.4          & 81.6          \\
\checkmark        &          &         &         & 48.6          & 76.2          & 84.6          \\
\checkmark        & \checkmark        &         &         & 49.5          & 76.3          & 84.9          \\
\checkmark        & \checkmark        & \checkmark       &         & 50.4          & 76.1          & 84.6          \\
\checkmark        & \checkmark        & \checkmark       & \checkmark       & \textbf{51.0} & \textbf{76.4} & \textbf{85.2} \\ \bottomrule[1.5pt]
\end{tabular}
}
\vspace{-0.5em}
\caption{Ablation study results on MSR-VTT. NM, CMI, QAF, and CVH denote the modules for narration matching, cross-modal interaction, query-aware filtering, and cross-view hard negative loss.}
\vspace{-1.0em}
\label{tab:tab3}
\end{table}

\begin{table}
\centering
\resizebox{0.8\columnwidth}{!}{%
\begin{tabular}{@{}c|c|ccc@{}}
\toprule[1.5pt]
Method                                                                       & Hyperparameter  & R@1           & R@5           & R@10          \\ \midrule
None                                                                            & -               & 48.1          & 74.7          & 82.7          \\ \midrule
\multirow{2}{*}{Top-k}                                                        & k = 3           & 50.3          & 76.4          & 84.9          \\
                                                                              & k = 4           & 50.2          & 76.4          & 85.0          \\ \midrule
\multirow{2}{*}{\begin{tabular}[c]{@{}c@{}}Nucleus \\ filtering\end{tabular}} & \(p\)  = 0.3 (2.55) & 50.3          & \textbf{76.6} & 84.5          \\
                                                                              & \(p\) = 0.4 (3.57)  & \textbf{51.0} & 76.4          & \textbf{85.2} \\ \bottomrule[1.5pt]
\end{tabular}
}
\vspace{-0.5em}
\caption{Change of accuracies depending on the filtering methods for MSR-VTT. The values in parentheses are the average numbers of frames selected during inference.}
\vspace{-1.0em}
\label{tab:filtering}
\end{table}

\begin{table}
\centering
\resizebox{0.7\columnwidth}{!}{%
\begin{tabular}{@{}c|ccccc@{}}
\toprule[1.5pt]
\( \alpha \)   & R@1           & R@5           & R@10          & MnR          & MdR           \\ \midrule
0   & 50.4          & 76.1          & 84.6          & 1.0          & \textbf{11.3} \\
0.5 & 50.8          & 76.4          & 84.8          & 1.0          & 11.4          \\
1.0 & \textbf{51.0} & \textbf{76.4} & \textbf{85.2} & \textbf{1.0} & 11.6          \\
1.5 & 50.1          & 76.3          & 85.2          & 1.0          & 11.4          \\ \bottomrule[1.5pt]
\end{tabular}
}
\vspace{-0.5em}
\caption{Performance changes depending on the hyperparameter of hard negative loss on MSR-VTT.}
\vspace{-1.0em}
\label{tab:cvh_alpha}
\end{table}

\subsection{Ablation Study}

In this section, we conduct an ablation study on MSR-VTT to thoroughly evaluate the effectiveness of the four modules proposed for narration utilization.

\textbf{Narration matching.} We first validate the impact of incorporating query-narration matching into the model. As shown in \cref{tab:tab3}, the R@1 increased by 4.1\%, demonstrating that the rich information in the narration serves as complementary data during the matching process. However, it is slightly lower than that of Cap4Video, indicating that the simple incorporation of frame-level captions is insufficient to fully leverage its potential. This also supports the need for comprehensive utilization through additional modules.

\textbf{Video-narration cross-modal interaction.} Beyond the complementary information provided by narration during the matching stage, we analyze the impact of feature enhancement through co-attention and temporal blocks. As shown in \cref{tab:tab3}, the use of enhanced features improves retrieval performance across all recall metrics. 
It indicates the importance of cross-modal interaction and the temporal information in video and narration features.

\textbf{Query-aware adaptive filtering.} Although the enhanced features contain rich information, not all features are relevant to the given query, and some erroneous information may be included. Therefore, proper filtering is necessary to ensure relevance and accuracy. The results in \cref{tab:tab3} demonstrate the importance of filtering, showing improved R@1 as irrelevant or inaccurate information is effectively removed. Furthermore, \cref{tab:filtering} highlights that unfiltered information can lead to a decline in performance, emphasizing the necessity of filtering methods, as significant performance gains are observed regardless of the filtering method employed. However, the optimal number of frames may vary depending on the samples. Notably, adaptive nucleus filtering with \( p=0.4 \) achieves higher R@1 accuracy while maintaining a similar average number of selected frames (3.57), compared to fixed selection with \( \text{k}=3 \) or \( \text{k}=4 \) used in previous studies. This supports the conclusion that adaptive nucleus filtering is more efficient than the fixed top-k selection approach.

\textbf{Cross-view hard negative loss.} The narration is also used to define a hard negative loss, including intra-modal similarity. The bottom row of \cref{tab:tab3} shows an improvement across all recall metrics, indicating that the proposed hard negative loss helps the model learn more discriminative features. \cref{tab:cvh_alpha} shows the dependence on the hyperparameter \( \alpha \), which controls the proportion of the hard negative loss term. The best performance is achieved with \( \alpha=1 \), implying the necessity of a substantial proportion of hard negatives for optimal performance.

\begin{figure}
    \centering
    \includegraphics[width=\linewidth]{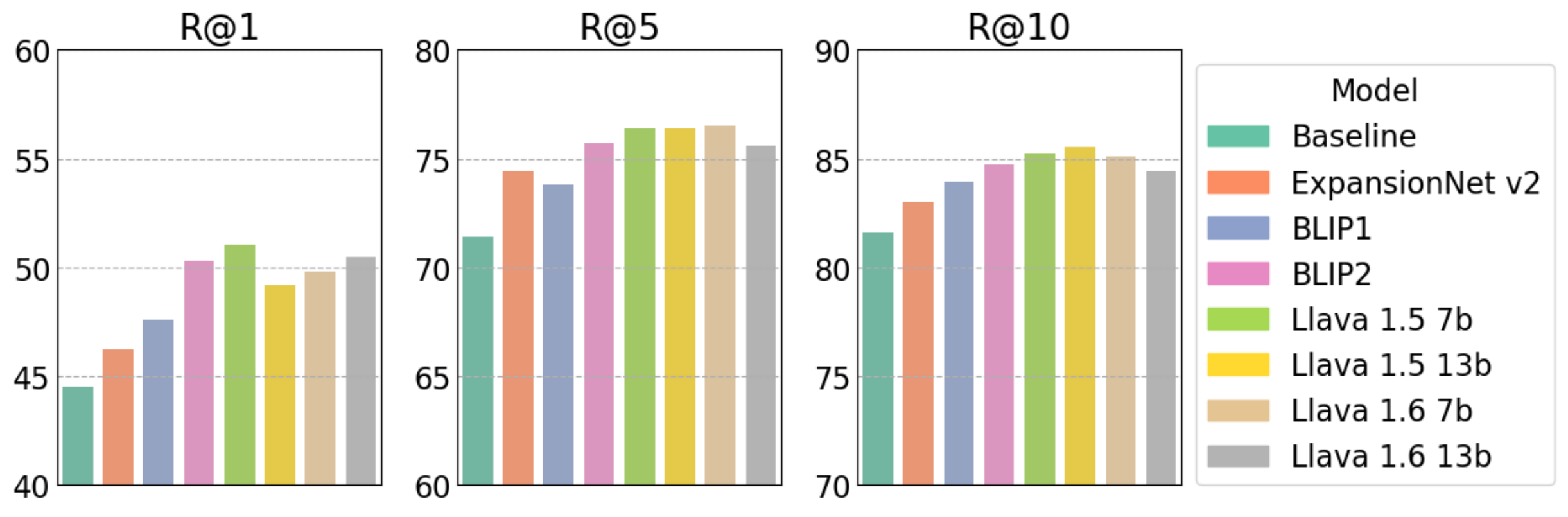}
    \vspace{-1.8em}
    \caption{Effectiveness of various captioners on MSR-VTT dataset. Note that except for the changes in the captioners, all other architecture is the same.}
    \vspace{-0.5em}
    \label{fig:captioner}
\end{figure}

\textbf{Impact of different narration generation models.} We analyze how different caption generation methods affect retrieval performance on NarVid. Specifically, as shown in \cref{fig:captioner}, we compared narrations from the output of several types of captioners: a traditional image-captioning model \cite{expansionnetv2}, popular vision-language pre-trained models \cite{blip,blip2}, and powerful large multimodal models \cite{llava, improvedllava}. The results show that the use of narration improves performance on all metrics regardless of the captioner, demonstrating that the strategy we propose works for a variety of caption generation models. Interestingly, we observed that more powerful VLMs generate detailed captions containing many query-irrelevant words. These well-generated but query-misaligned captions could negatively impact retrieval performance. Detailed experimental results about the caption generation can be found in Supplementary \cref{sec:supple:captioner} %Sec. C

%-------------------------------------------------------------------------

\subsection{Qualitative Results}

\begin{figure}
    \centering
    \includegraphics[width=\linewidth]{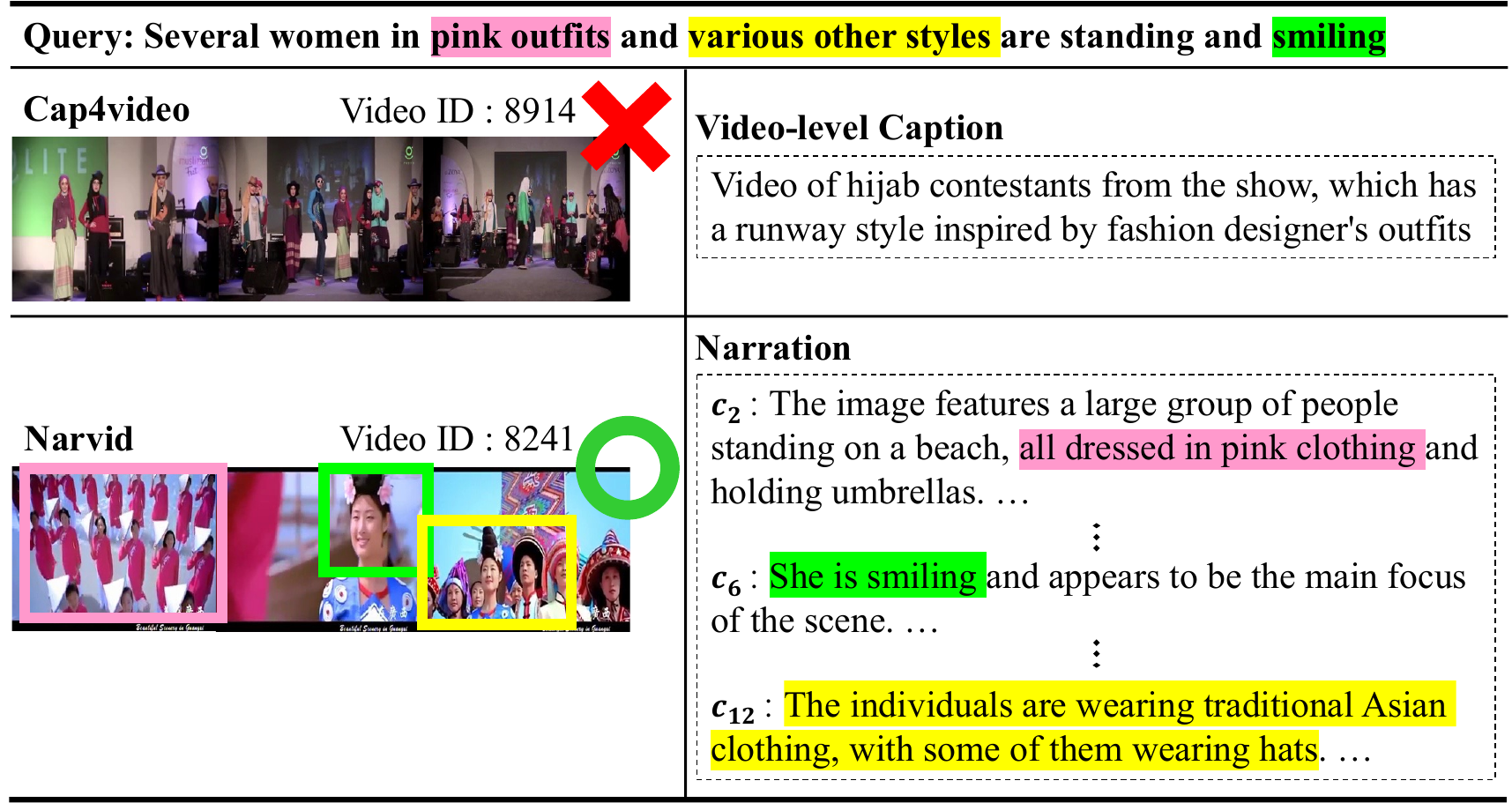}
    \vspace{-0.5em}
    \caption{Text-to-video retrieval results (R@1) on MSR-VTT test set. For the Cap4Video, the video-level caption does not match the words in the query. For NarVid, narration contains richer information over frames, and it leads to the correct retrieval results.}
    \vspace{-1em}
    \label{fig:visualization}
\end{figure}

We also compare the actual retrieval result (R@1) from our NarVid with the Cap4Video \cite{cap4video} using a video-level caption. The video-level caption summarizing the entire video struggles to reflect the whole content of the video, especially in videos with frequent scene changes. In contrast, the frame-level captions not only better capture the whole video contents but also help focus on the segments most relevant to the query. As shown in \cref{fig:visualization}, the video-level caption prioritizes the theme of outfits, leading to incorrect retrieval results linked to fashion shows. Conversely, our narration provides rich information, including outfits, diverse styles, and smiling expressions. Utilizing this information properly, NarVid could obtain more accurate results.

\section{Conclusions}
\label{sec:conclusion}

We propose the NarVid framework which focuses on the comprehensive utilization of frame-level generated captions, named narration, for the text-to-video retrieval task. The narration naturally provides rich semantic information arranged in chronological order. We utilize narrations in strategic ways, such as feature enhancement through cross-modal interaction, filtering out inappropriate information, and defining a novel similarity and loss function for matching and training. Extensive experiments on benchmark datasets validate the superiority of NarVid and reveal its potential effectiveness for retrieval tasks.

\textbf{Limitations.} To improve real-world applicability, a pre-computation and storage process for narration generation is necessary for the NarVid framework. Efficient narration generation techniques could be employed to reduce the pre-computation burden. In addition, although we demonstrated that our framework works over different caption generators in \cref{fig:captioner}, the improvement in retrieval performance is closely tied to the capability of the caption generator. A more thorough investigation into the caption generation process, incorporating prompt engineering and advanced generation techniques, could further enhance the retrieval performance.
%-------------------------------------------------------------------------

%\newpage
\section*{Acknowledgments}
\label{sec:acknowledgments}
This work was supported in part by Institute of Information \& communications Technology Planning \& Evaluation (IITP) grant funded by the Korea government(MSIT) (No. RS-2021-II212068, Artificial Intelligence Innovation Hub) in part by the MSIT(Ministry of Science and ICT), Korea, under the ITRC(Information Technology Research Center) support program(IITP-2024-RS-2024-00437718) supervised by the IITP(Institute for Information \& Communications Technology Planning \& Evaluation), and in part by Institute of Information \& communications Technology Planning \& Evaluation (IITP) grant funded by the Korea government(MSIT) (No.RS-2023-00227431, Development of 3D space digital media standard technology).
\nocite{*} 
{
    \small
    \bibliographystyle{ieeenat_fullname}
    \bibliography{main}
}

\clearpage
\maketitlesupplementary

% 섹션 카운터 초기화
\setcounter{section}{0} % 섹션 카운터를 0으로 설정

% figure, table, equation 본문에 이어서 (본문과 같이 컴파일 할 때는 주석처리 필요)
\setcounter{equation}{11}
\setcounter{figure}{4}
\setcounter{table}{5}

% 섹션 번호 형식을 A, B, C로 변경
\renewcommand{\thesection}{\Alph{section}}

\noindent The supplementary is organized as follows:

\begin{itemize}
\item The details of the multi-grained matching method used for query-video-narration alignment in \cref{sec:supple:multi-granularity}.
\item Further information on the datasets used in the experiments, along with additional implementation details, in \cref{sec:supple:experimental-settings}.
\item The details of the frame-level caption generation process in \cref{sec:supple:captioner}.
\item Additional ablation studies in \cref{sec:supple:ablation}.
\item Further qualitative analyses for each dataset in \cref{sec:supple:qualitative}.
\end{itemize}

\section{Multi-Granularity Matching Module}
\label{sec:supple:multi-granularity}

Inspired by previous research \cite{cap4video}, we propose a modified multi-granularity matching process with our proposed narration and the result of a query-aware adaptive filtering module.

\begin{figure}[t]
    \centering
    \includegraphics[width=\linewidth]{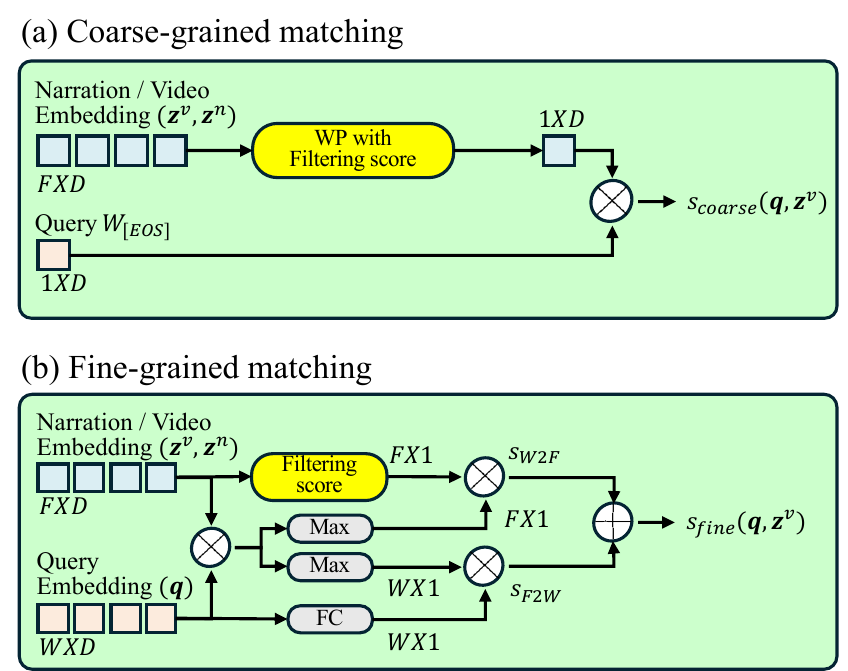}
    \caption{Process of coarse-grained and fine-grain matching. In the yellow box, we utilize the results of nucleus filtering for weight values.}
    \label{fig:multi-granularity}
\end{figure}

\textbf{Coarse-grained matching.} In \cref{fig:multi-granularity} (a), we illustrate the coarse-grained matching process. For both video and narration, the sequence of embedding vectors for frames and captions is transformed into a single vector through weighted pooling. The cosine similarity between this pooled vector and the \( w_{[EOS]} \) vector of the query is then computed to obtain the coarse matching scores \( s_{coarse}(\boldsymbol{q}, \boldsymbol{z}^v) \) and \( s_{coarse}(\boldsymbol{q}, \boldsymbol{z}^n) \). Here, it should be remarked that the weight values for the pooling are determined by the query-aware filtering module. In the query-aware adaptive filtering module described in \cref{sec:method:filtering}, we evaluate individual frames and captions based on their similarities to the given query, and obtain the final filtering scores for the selected frames \( \boldsymbol{z}^v \) and captions \( \boldsymbol{z}^n \) through filtering and softmax. An example of the filtering scores is shown in \cref{fig:framework} (b) %Fig. 2 (b). 

By using the filtering scores as weights for pooling, we can emphasize the most relevant frames in the query-video matching process and the most relevant captions in the query-narration matching process.

\textbf{Fine-grained matching.} \cref{fig:multi-granularity} (b) shows the details of the proposed fine-granularity matching process. For query-video matching, we first compute a similarity matrix with the cosine similarity of all possible pairs of frame embeddings and word embeddings and take frame-wise and word-wise maximums to get two vectors of maximum similarity values. The frame-wise maximum similarity vector is then used to compute the inner product with the filtering score to obtain the similarity score \(s_{W2F}\). We also compute a similarity score \(s_{F2W}\) using a word-wise maximum similarity vector and a weight vector from an FC layer that is trained in an end-to-end manner. Finally, the two similarity scores are summed to obtain a fine-granularity score \( s_{fine}(\boldsymbol{q}, \boldsymbol{z}^v) \). For query-narration matching, the same process is performed with narration embedding vectors to obtain the score \( s_{fine}(\boldsymbol{q}, \boldsymbol{z}^n) \).

\section{Experimental Settings}
\label{sec:supple:experimental-settings}

\subsection{Datasets}
To validate our model, we utilized the MSR-VTT \cite{msrvtt}, MSVD \cite{msvd}, VATEX \cite{vatex}, and DiDeMo \cite{didemo} datasets, which are commonly employed in previous studies. \cref{tab:dataset_statistics} shows a summary of the properties of four datasets.

MSR-VTT provides 10K web video clips, totaling 38.7 hours of content, along with 200K clip-sentence pairs. These clips, collected from the YouTube platform, cover a wide range of visual content across various comprehensive categories. Each clip ranges from 10 to 30 seconds in length and is annotated with approximately 20 natural sentences generated by 1,327 Amazon Mechanical Turk (AMT) workers. MSR-VTT is typically divided into three types of splits (training/testing): full split (7K/3K), 1k-A split (9K/1K) \cite{jsfusion}, and 1k-B split (6.5K/1K) \cite{1kb-split}. Among these, the 1k-A split is most commonly used for performance comparison, and we adopted this in our evaluation.

The MSVD dataset consists of over 85,000 English descriptions for 2K video snippets. These descriptions were generated by AMT workers who summarized the action in each short video snippet with a single sentence. The video clips are generally set to a short length of 4 to 10 seconds, each aiming to depict a clear event or action. 

VATEX comprises over 35K videos, each averaging 10 seconds in length, accompanied by 700K captions in both Chinese and English. This dataset includes more than 206,000 parallel English-Chinese translations, featuring long sentences with relatively diverse lexical characteristics.

DiDeMo consists of videos sourced from Flickr, which are trimmed to a maximum length of 30 seconds and divided into 5-second segments to reduce annotation complexity. The dataset is split into training, validation, and test sets containing 8,395, 1,065, and 1,004 videos, respectively. In total, the dataset includes 26,892 moments, with each moment potentially associated with multiple descriptions by different annotators. Following the approach of previous studies \cite{clip4clip, cap4video}, we concatenated all sentence descriptions of each video into a single sentence query for evaluation purposes. 

\begin{table}[]
\centering
\begin{tabular}{@{}c|cccc@{}}
\toprule[1.5pt]
Dataset       & Videos & Captions & Length & Source                                                       \\ \midrule
MSR-VTT  \cite{msrvtt} & 10K             & 200K              & 10-30s          & Youtube                                                               \\ \midrule
MSVD \cite{msvd}       & 2K              & 8.5K              & 4-10s           & Youtube                                                               \\ \midrule
VATEX \cite{vatex}     & 35K             & 700K              & Avg 10s         & \begin{tabular}[c]{@{}c@{}}Kinetics\\ -600,   \\ Youtube\end{tabular} \\ \midrule
DiDeMo \cite{didemo}   & 10K             & -                 & Max 30s         & Flickr                                                                \\ \bottomrule[1.5pt]
\end{tabular}
\caption{Statistics of four benchmark datasets.}
\label{tab:dataset_statistics}
\end{table}

\subsection{Implementation Details}
\label{sec:supple:experimental-settings:implementation_details}
To ensure reproducibility, we provide additional implementation details in this section. Following the settings of CLIP4Clip \cite{clip4clip}, we configure the initial learning rate to 1e-7 for the CLIP \cite{clip} module, while other modules are set to 1e-4. The Adam optimizer is utilized, accompanied by a cosine scheduling strategy and a warm-up proportion of 0.1. Additionally, the temperature hyperparameter for all softmax functions is set to 0.1.

For preprocessing all datasets, we follow the data preparing process of CLIP4Clip. Also, we uniformly sample each video data at 12 frames for MSR-VTT, MSVD, and VATEX datasets, and sample 32 frames for DiDeMo to cover longer video duration. 

For the hard negative loss, the hyperparameter settings vary by dataset. For MSR-VTT with the ViT-B/32 backbone, we set \(\lambda = 0.7\), \(\eta = 1.8\), and \(\alpha = 1\).
In contrast, for the ViT-B/16 backbone, which demonstrates better performance, we focused on hard negative selection and loss weighting for performance optimization.
Accordingly, for MSR-VTT, we set \(\lambda = 1.1\), \(\eta = 2.0\), and \(\alpha = 2\). For MSVD, the values are \(\lambda = 0.9\), \(\eta = 1.8\), and \(\alpha = 1\). For VATEX, we use \(\lambda = 0.9\), \(\eta = 1.8\), and \(\alpha = 0.8\). For DiDeMo, we set \(\lambda = 1.0\), \(\eta = 1.9\), and \(\alpha = 1\).
All experiments are conducted on two NVIDIA RTX A6000 GPUs to ensure consistency in our experimental setup.

\begin{table}[t]
\centering
\begin{tabular}{@{}c|l@{}}
\toprule[1.5pt]
Type      & \multicolumn{1}{c}{Prompt \& Characteristic}\\ \midrule
                   & {\ul Usage Prompt:} \\
\multirow{2}{*}{A} & \textit{\begin{tabular}[c]{@{}l@{}}Briefly describe the object in the image in one \\ sentence.\end{tabular}}\\
                   & \\
                   & Characteristic: Simple, Single sentence \\ \midrule
                   & {\ul Usage Prompt:}\\
\multirow{2}{*}{B} & \textit{\begin{tabular}[c]{@{}l@{}}You want to view an image, one of the frames in\\ a video clip, and organize the text description\\ into a single sentence. Follow these steps. 1. \\ Analyze the input image by dividing it into \\ objects, actions, and other parts. 2. Create a\\ one-sentence text description based on your\\ previous analysis. 3. Output only the processed\\ text without any additional description.\end{tabular}} \\
                   & \\
                   & Characteristic: Structured, Single sentence\\ \midrule
                   & {\ul Usage Prompt:}\\
\multirow{2}{*}{C} & \textit{\begin{tabular}[c]{@{}l@{}}Please describe this image for image-captioning \\ task\end{tabular}}\\
                   &\\
                   & Characteristic: Simple, Not single sentence\\ \bottomrule[1.5pt]
\end{tabular}
\caption{Three different types of prompts used for caption generation. The characteristic of each type is also denoted.}
\label{tab:prompt_char}
\end{table}

\begin{table}[t]
\centering
\resizebox{\columnwidth}{!}{ % 너비를 원컬럼에 맞춤
\begin{minipage}{\columnwidth}
\fontsize{10}{\baselineskip}\selectfont % 전체 글자 크기 설정
\setlength{\tabcolsep}{19pt} % 열 간격 설정 (12pt로 조정)
\begin{tabular}{@{}cccc@{}}
\toprule[1.5pt]
\multirow{2}{*}{Prompt 
type} & \multicolumn{3}{c}{Text-to-video retrieval} \\ \cmidrule(l){2-4} 
                                      & R@1              & R@5             & R@10            \\ \midrule
\multicolumn{1}{c|}{A}                & 52.6             & \textbf{82.0}   & \textbf{89.5}   \\
\multicolumn{1}{c|}{B}                & 52.6             & 81.6            & 89.0            \\
\multicolumn{1}{c|}{C}                & \textbf{53.1}    & 81.4            & 88.8            \\ \bottomrule[1.5pt]
\end{tabular}
\end{minipage}
}
\caption{Text-to-video retrieval results on MSVD depend on the types of prompts used for caption generation.}
\label{tab:prompt_results}
\end{table}

\begin{figure*}[]
  \centering
  \includegraphics[width=1.0\textwidth]{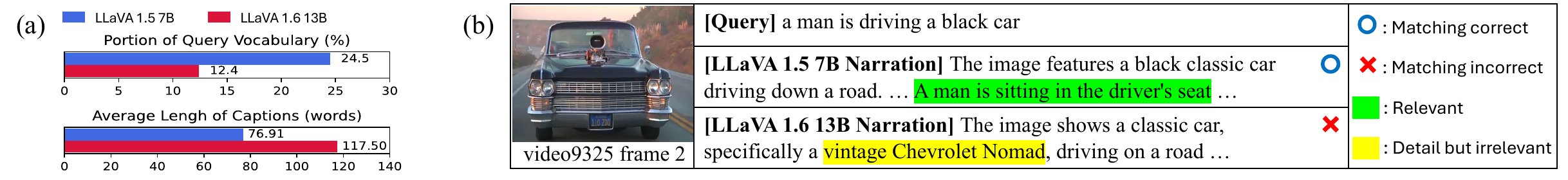}
   \caption{Narration comparison on MSR-VTT across different LLaVA versions.}
   \label{fig:llava_visualization}
\end{figure*}

\section{Caption Generator}
\label{sec:supple:captioner}
\subsection{Prompt Strategies}
\label{sec:supple:captioner:prompt}
When using large multimodal models (LMMs) for caption generation, both the caption quality and inference time can vary depending on the prompt. Although our NarVid framework generates narrations offline, the computational resources and time required for pre-generating captions remain practical challenges in our experiments. Finding the optimal prompt for each dataset requires substantial experimentation, which may be beyond the scope of our current research. To address this challenge, we utilized the MSVD dataset \cite{msvd} as a testbed, which requires the least computational resources for inference among the datasets. We first evaluated the retrieval performance across various prompts on MSVD and then applied our findings to other datasets.

\begin{table}[]
\centering
\resizebox{\columnwidth}{!}{%
\begin{minipage}{\columnwidth}
\fontsize{10}{\baselineskip}\selectfont % 전체 글자 크기 설정
\setlength{\tabcolsep}{11pt} % 열 간격 설정 (12pt로 조정)
\begin{tabular}{@{}rcccc@{}}
\toprule[1.5pt]
\multirow{2}{*}{Captioner} & \multirow{2}{*}{Time(s)} & \multicolumn{3}{c}{Text-to-video retrieval}   \\ \cmidrule(l){3-5} 
                           &                          & R@1           & R@5           & R@10          \\ \midrule
LLaVA 0.5B                 & \textbf{7.92}            & 50.0          & 76.2          & 84.9          \\
LLaVA 7B                   & 25.31                    & \textbf{51.0} & \textbf{76.4} & \textbf{85.2} \\ \bottomrule[1.5pt]
\end{tabular}%
\end{minipage}
}
\caption{Comparison of text-to-video retrieval results on MSR-VTT based on the LLaVA model size.}
\label{tab:effective_caption_generator}
\end{table}

As shown in \cref{tab:prompt_char}, we provide several examples of prompts with their specific characteristics. We used the LLaVa 1.5 7B model \cite{llava} as a baseline model for the experiments. Prompt A is a simple request to describe the objects within each frame in a single sentence. Prompt B is more structured, requiring a single-sentence description that includes objects, actions, and other parts. Prompt C removes the single-sentence constraint, allowing for a more detailed description of each frame. As shown in \cref{tab:prompt_results}, prompt C achieves the highest performance in R@1, and we use it for other experiments. Given the apparent potential for improvement in these captions, we expect that various advanced inference techniques, such as in-context learning, will enhance the performance in future work. The captions used for experiments will be publicly available with our code.
\subsection{Time Efficiency for Caption Generation}
In our experiments, we generated all frame-level captions per video offline. As shown in \cref{tab:effective_caption_generator}, we mainly use the basic LLaVa 1.5 7B model as a captions generator, therefore the inference time can be reduced with inference boosting toolkits. One interesting point is that a lighter frame captioner (0.5B) can achieve considerable performance with less computing time. 

\subsection{Performance Discrepancy by LLaVA 1.5 7B vs 1.6 13B}
In general, the larger VLM tends to generate more detail and long captions, but all of them are not necessarily related to given queries, which shows less portion of query vocabulary (\cref{fig:llava_visualization} (a)). The well-generated captions with query-irrelevant words may cause a negative effect on the retrieval performance, as shown in \cref{fig:llava_visualization} (b).

\section{Ablation Study}
\label{sec:supple:ablation}
\begin{table}[t]
\centering
\begin{tabular}{@{}ccccrr@{}}
\toprule[1.5pt]
\multirow{2}{*}{Matrix}                                & \multicolumn{5}{c}{Text-to-video retrieval (zero-shot)}                                   \\ \cmidrule(l){2-6} 
 & R@1           & R@5           & R@10          & MdR          & MnR           \\ \midrule
\multicolumn{1}{c|}{\(\boldsymbol{S}_{qv}\)}                     & \textbf{31.4} & \textbf{53.8} & \textbf{62.5} & \textbf{4.0} & \textbf{41.3} \\
\multicolumn{1}{c|}{\(\boldsymbol{S}_{qn}\)}                     & 13.7          & 25.6          & 31.5          & 64.5         & 188.9         \\
\multicolumn{1}{c|}{\(\boldsymbol{S}_{sum}\)} & 21.5          & 37.3          & 43.8          & 15.0         & 109.9         \\
\multicolumn{1}{c|}{\(\boldsymbol{S}_{fusion}\)}                 & 27.3          & 47.0          & 56.1          & 6.0          & 63.9          \\ \bottomrule[1.5pt]
\end{tabular}
\caption{Results depending on similarity matrices on MSR-VTT for zero-shot retrieval evaluation. Setting: we extracted CLIP features from the videos and their narrations to compare with the queries' CLIP features without any training. \(\boldsymbol{S}_{qv}\) and \(\boldsymbol{S}_{qn}\) are the query-video and query-narration similarity matrices, each. \(\boldsymbol{S}_{sum}\) indicates the element-wise sum of the two matrices, while \(\boldsymbol{S}_{fusion}\) represents the element-wise sum of the two normalized matrices.}
\label{tab:zeroshot_msrvtt}
\end{table}

\begin{table}[]
\centering
\resizebox{\columnwidth}{!}{%
\begin{tabular}{@{}ccccccc@{}}
\toprule[1.5pt]
\multirow{3}{*}{Matrix}                          & \multicolumn{6}{c}{Text-to-video retrieval}             \\ \cmidrule(l){2-7} 
& \multicolumn{3}{c}{MSR-VTT}                   & \multicolumn{3}{c}{DiDeMo}                    \\ \cmidrule(l){2-7} 
& R@1           & R@5           & R@10          & R@1           & R@5           & R@10          \\ \midrule
\multicolumn{1}{c|}{\(\boldsymbol{S}_{qv}\)}     & 46.3          & 74.3          & 82.9          & 48.9          & 77.7          & 84.8          \\
\multicolumn{1}{c|}{\(\boldsymbol{S}_{qn}\)}     & 46.0          & 75.2          & 83.2          & 46.7          & 72.9          & 82.2          \\
\multicolumn{1}{c|}{\(\boldsymbol{S}_{sum}\)}    & 51.0          & 76.4          & 85.2          & 52.5          & 79.1          & 85.7          \\
\multicolumn{1}{c|}{\(\boldsymbol{S}_{fusion}\)} & \textbf{51.0} & \textbf{76.4} & \textbf{85.2} & \textbf{53.0} & \textbf{79.5} & \textbf{86.0} \\ \bottomrule[1.5pt]
\end{tabular}%
}
\caption{Retrieval performance comparison based on the similarity matrices on our NarVid framework. Note that \(\boldsymbol{S}_{qv}\), \(\boldsymbol{S}_{qn}\) \(\boldsymbol{S}_{sum}\) and \(\boldsymbol{S}_{fusion}\) are same operation as \cref{tab:zeroshot_msrvtt}.}
\label{tab:compare_fusion}
\end{table}

\subsection{Effectiveness of Narration Utilization}
\begin{figure}
    \centering
    \includegraphics[width=\linewidth]{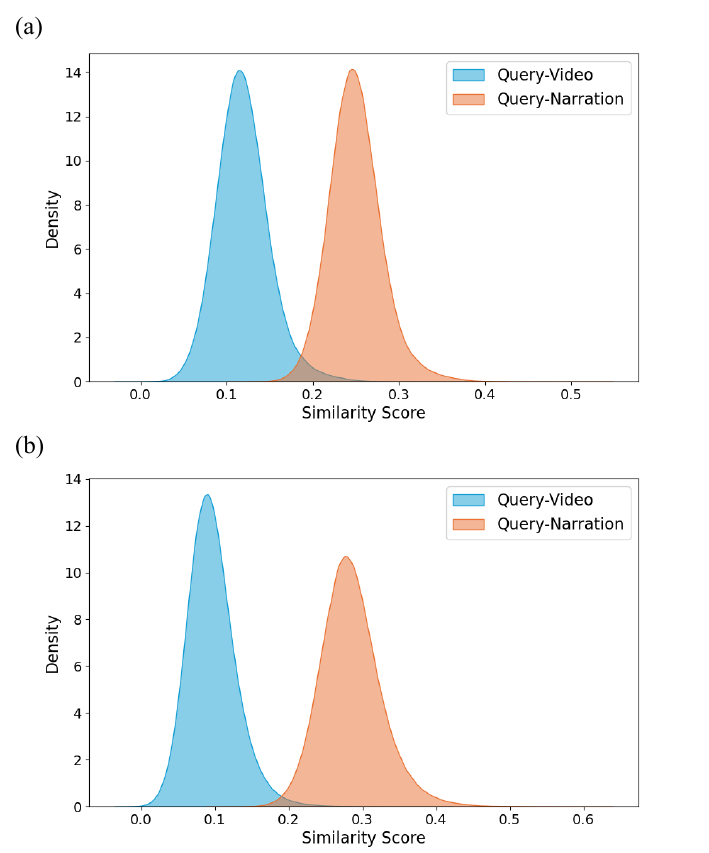}
    \caption{Visualization of similarity matrix distribution for query-video and query-narration. (a) results for MSR-VTT. (b) results for DiDeMo}
    \label{fig:sim_distribution}
\end{figure}
\textbf{Zero-shot retrieval performance.}
\textit{Can captions from good models such as LMMs deliver outstanding results on their own?}
To answer this question, in \cref{tab:zeroshot_msrvtt}, we evaluate zero-shot retrieval to verify the importance of effectively utilizing narration information without training. For the experiment setting, we extracted CLIP features from each video and its narration, applied mean pooling, and evaluated zero-shot retrieval performance with given queries. In the table, the query-to-video retrieval shows reasonable retrieval performance without training, while the query-to-narration retrieval shows significantly lower performance. Although an element-wise matrix summation or the fusion approach used in our NarVid framework shows some improvements, the retrieval performances remain insufficient for practical retrieval scenarios. These experimental results once again support the claim of our framework that the information from the generative model should be utilized appropriately, including training.

\textbf{Similarity matching fusion.}
After the training process on our  NarVid, we utilize the query-video similarity matrix \(\boldsymbol{S}_{qv}\) as well as the additional query-narration similarity matrix \(\boldsymbol{S}_{qn}\) during the inference phase. However, as shown in \cref{fig:sim_distribution}, the distributions of the elements in the two matrices are notably different. Therefore, a simple summation of these matrices could lead to an overemphasis on one aspect over the other. To mitigate this issue, we standardize the element values in each matrix separately before summation, using their means (\(\mu^{qv}, \mu^{qn}\)) and standard deviations (\(\sigma^{qv}, \sigma^{qn}\)). The two standardized matrices are then fused to form the final score matrix \(\boldsymbol{S}_{fusion}\), ensuring a balanced consideration of both aspects, which is defined in \cref{eq:infer}. %Eq. (11).

\cref{tab:compare_fusion} presents the effect of the similarity matrix fusion on the MSR-VTT \cite{msrvtt} and DiDeMo \cite{didemo} dataset. The improvement of over 4.4\% in R@1 performance when combining the two matrices indicates that the information from video and narration works complementarily. Furthermore, the results on DiDeMo confirm that \(\boldsymbol{S}_{fusion}\) outperforms the element-wise summation. The results demonstrate the effectiveness and robustness of our fusion method, especially when the distributions of the two matrices differ regarding their standard deviations, as illustrated in \cref{fig:sim_distribution} (b).

\begin{table}[]
\centering
\resizebox{\columnwidth}{!}{%
\begin{minipage}{\columnwidth}
\fontsize{10}{\baselineskip}\selectfont % 전체 글자 크기 설정
\setlength{\tabcolsep}{11pt} % 열 간격 설정 (12pt로 조정)
\begin{tabular}{@{}ccccc@{}}
\toprule[1.5pt]
\multicolumn{2}{c}{Temporal modeling}                                 & \multicolumn{3}{c}{Text-to-video retrieval} \\ \midrule
\(\phi_{temp}(\boldsymbol{\hat{v}})\) & \(\phi_{temp}(\boldsymbol{\hat{n}})\) & R@1       & R@5       & R@10      \\ \midrule
                        & \multicolumn{1}{c|}{\checkmark}             & 49.1      & 75.9      & 84.8      \\
\checkmark                                   & \multicolumn{1}{c|}{}   & 50.2      & 75.8      & 85.4      \\
\checkmark                        & \multicolumn{1}{c|}{\checkmark}   & 51.0      & 76.4      & 85.2      \\ \bottomrule[1.5pt]
\end{tabular}%
\end{minipage}
}
\caption{Comparison of the impact of temporal modeling on performance across modalities on MSR-VTT}
\label{tab:temporal_modeling}
\end{table}

\subsection{Temporal Modeling Details.}
The temporal block in \cref{eq:temporal} is defined as \(\phi_{temp}(\boldsymbol{\hat{v}}) = Transformer(\boldsymbol{\hat{v}} + \boldsymbol{P}) + \boldsymbol{\hat{v}}\), where \(\boldsymbol{P}\) represents the positional embedding, as used in CLIP4Clip \cite{clip4clip}. Temporal modeling is commonly applied to videos, as supported by Row 1 of \cref{tab:temporal_modeling}. Similarly, although narrations have inherent temporal orders, we believe that effectively capturing the temporal relationships among frame-level captions enhances representation. Notably, removing \(\phi_{temp}\) from the narration led to a 0.8\% decrease in R@1, highlighting the significance of temporal modeling.

\subsection{Cross-View Hard Negative Loss}
\begin{figure}
    \centering
    \includegraphics[width=\linewidth]{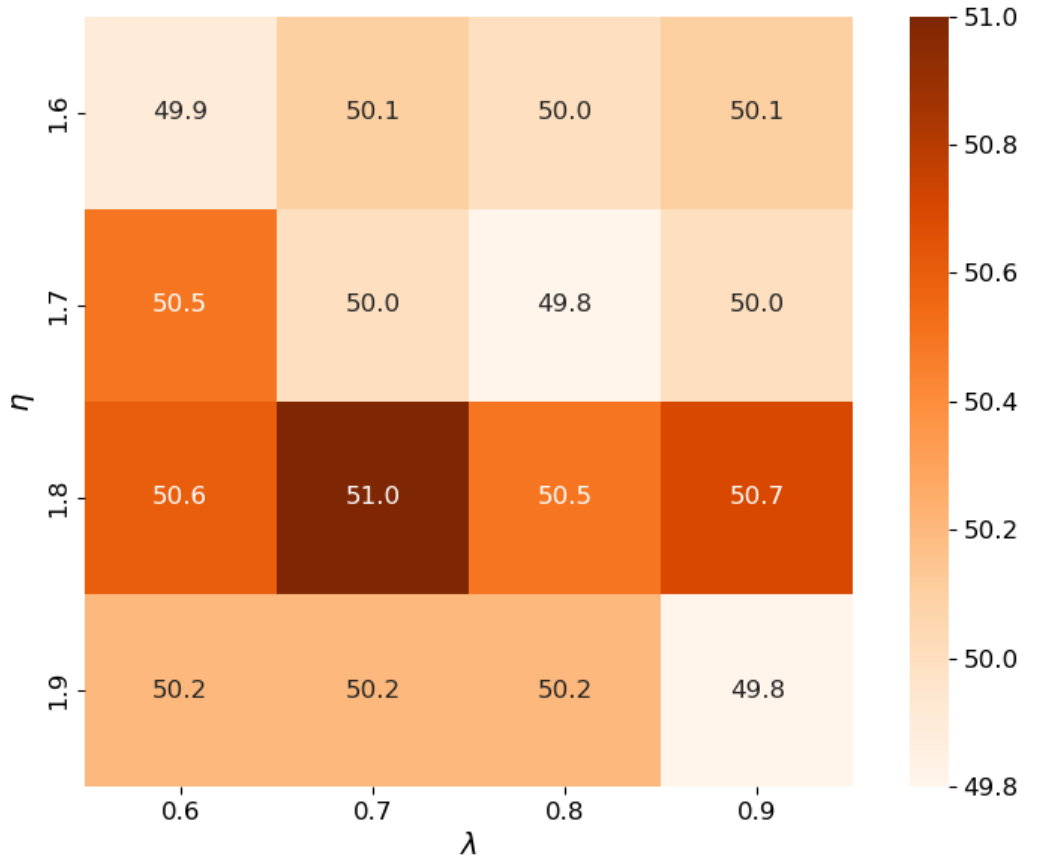}
    \caption{Ablation study results of \(\lambda\) and \(\eta\), used in the hard negative loss, with \(\alpha\) set to 1, conducted on the MSR-VTT.}
    \label{fig:cvh_hyperparameter}
\end{figure}
\cref{fig:cvh_hyperparameter} shows the performance of R@1 with varying values of \(\lambda\) and \(\eta\), which are hyperparameters used in the hard negative rank loss of \cref{eq:hard_selection,eq:hard_rank_loss}. %Eqs. (6) and (8).

\(\lambda\) is a scaling factor that establishes the threshold for selecting hard negatives within the query-video and query-narration similarity matrices. Likewise, \(\eta\) serves as a scaling factor that modifies the thresholds in the cross-view hard negative loss. The balance between \(\lambda\) and \(\eta\) is crucial for optimal performance, achieving its best results at \(\lambda=0.7\) and \(\eta=1.8\).

\begin{figure}[]
  \centering
  \includegraphics[width=1.0\linewidth]{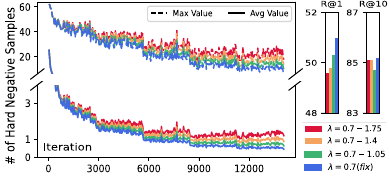}
   \caption{Variations in selected hard negatives with the increase of \(\lambda\), along with the corresponding recall results on MSR-VTT.}
   \label{fig:lambda_ablation}
\end{figure}

\begin{figure}[]
  \centering
  \includegraphics[width=1.0\linewidth]{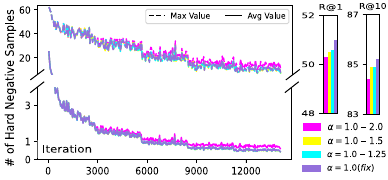}
   \caption{Variations in selected hard negatives with the increase of \(\alpha\), along with the corresponding recall results on MSR-VTT.}
   \label{fig:alpha_ablation}
\end{figure}

Intuitively, employing the same hyperparameters during training may lead to a decrease in the selection of hard negative samples, consequently diminishing the impact of the hard negative loss. This raises an intriguing question about the benefits of focusing on hard negative samples in the later stages of training.
To address this question, we linearly increased \(\lambda\) to acquire more hard negative samples during experimentation. As shown in \cref{fig:lambda_ablation}, our findings reveal that while this approach successfully increased the quantity of hard negative samples, it resulted in a degradation of performance.
This indicates that forcing easy samples to be treated as hard negatives may hinder representation learning.

\cref{fig:alpha_ablation} presents the results of linearly increasing \(\alpha\), which similarly shows performance degradation. Notably, in the case of increasing \(\alpha\) from 1.0 to 2.0, which emphasizes the hard negative loss, we found that the number of selected hard negative samples actually increased. The result suggests that dynamically modifying the hard negative loss weight to focus more on hard negatives may negatively impact the learned representations. \\

We conducted additional experiments with different types of hard negative loss functions; using the same unified hard negative samples \(H_i\) (for query-to-video and query-to-narration) and \(H_i^T\) (for video-to-query and narration-to-query), we evaluate the performance of hard negative loss based on the InfoNCE loss, \(L_{NCE}^H\), defined as follows:

\begin{align} 
    &L_{qv}^{H}=-\frac{1}{2B} \sum_{i=1}^B \left\{ \log \frac{e^{\boldsymbol{S}_{qv}(i, i)}}{\displaystyle\sum_{{j}\in{H_i}} e^{\boldsymbol{S}_{qv}(i, j)}} + \log \frac{e^{\boldsymbol{S}_{qv}(i, i)}}{\displaystyle\sum_{{j}\in{H_i^T}} e^{\boldsymbol{S}_{qv}(j, i)}} \right\}, \nonumber
\\
    &L_{qn}^{H}=-\frac{1}{2B} \sum_{i=1}^B \left\{ \log \frac{e^{\boldsymbol{S}_{qn}(i, i)}}{\displaystyle\sum_{{j}\in{H_i}} e^{\boldsymbol{S}_{qn}(i, j)}} + \log \frac{e^{\boldsymbol{S}_{qn}(i, i)}}{\displaystyle\sum_{{j}\in{H_i^T}} e^{\boldsymbol{S}_{qn}(j, i)}} \right\}, \nonumber
\\
        &L_{NCE}^{H} = \frac{1}{2} (L_{qv}^{H} + L_{qn}^{H}).
    \label{eq:infonce_hard}
\end{align}
\\
As shown in \cref{tab:hard_loss_comparison}, both types of hard negative loss assist the model in learning discriminative features. This highlights the effectiveness and robustness of our method that leverages unified negative samples from two different views: inter-modality (query-video) and intra-modality (query-narration).

\begin{table}[]
\centering
\begin{tabular}{@{}cccc@{}}
\toprule[1.5pt]
\multirow{2}{*}{Loss}             & \multicolumn{3}{c}{Text-to-video retrieval} \\ \cmidrule(l){2-4} 
                                           & R@1              & R@5             & R@10            \\ \midrule
\multicolumn{1}{c|}{Only \(L_{NCE}\)}      & 50.4             & 76.1            & 84.6            \\
\multicolumn{1}{c|}{\(L_{NCE}+L_{NCE}^H\)} & 50.9             & 76.2            & 85.2            \\
\multicolumn{1}{c|}{\(L_{NCE}+L_{CVH}\)}     & \textbf{51.0}    & \textbf{76.4}   & \textbf{85.2}   \\ \bottomrule[1.5pt]
\end{tabular}
\caption{Comparison of performance results for different loss configurations on MSR-VTT.}
\label{tab:hard_loss_comparison}
\end{table}

\subsection{Cost per Each Module.} For the number of trainable parameters, the two Cross-Modal Interactions have \(\phi_{co-attn}\) (3.7M) and \(\phi_{temp}\) (12.7M), respectively. The Narration Matching has 0.3M, leading to a 12\% increase over the baseline \cite{clip4clip} (136.2M). For FLOPs, the Narration Matching shows a 30.5G increase due to the expanded caption encoding, reaching 1.5 times the baseline (54.4G). The others have little impact.

\section{Qualtative Analysis}
\label{sec:supple:qualitative}
\begin{figure*}
    \centering
    \includegraphics[width=0.90\linewidth]{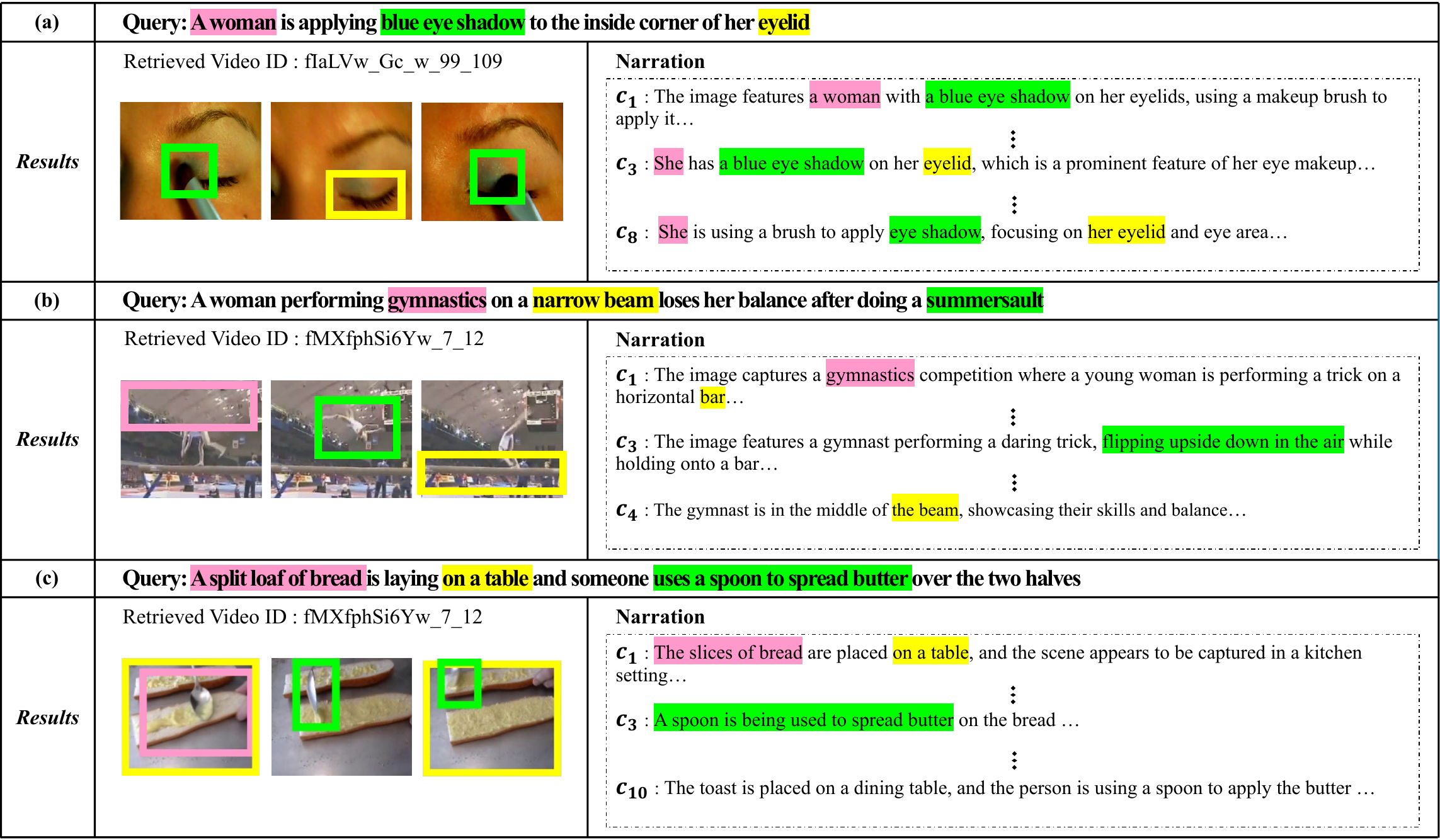}
    \caption{Text-to-video retrieval results on MSVD. (a) In the provided query, NarVid effectively identified the subject, a woman, and accurately recognized details such as the blue eye shadow color and the makeup situation on the eyelid. (b) NarVid demonstrates a good understanding of gymnastics, specifically the narrow beam and the somersault action, describing it as flipping upside down in detail. (c) NarVid not only understands the layout of the bread and table but also detects the spoon's movement to locate the right video.}
    \label{fig:msvd_visualization}
\end{figure*}

\begin{figure*}
    \centering
    \includegraphics[width=0.90\linewidth]{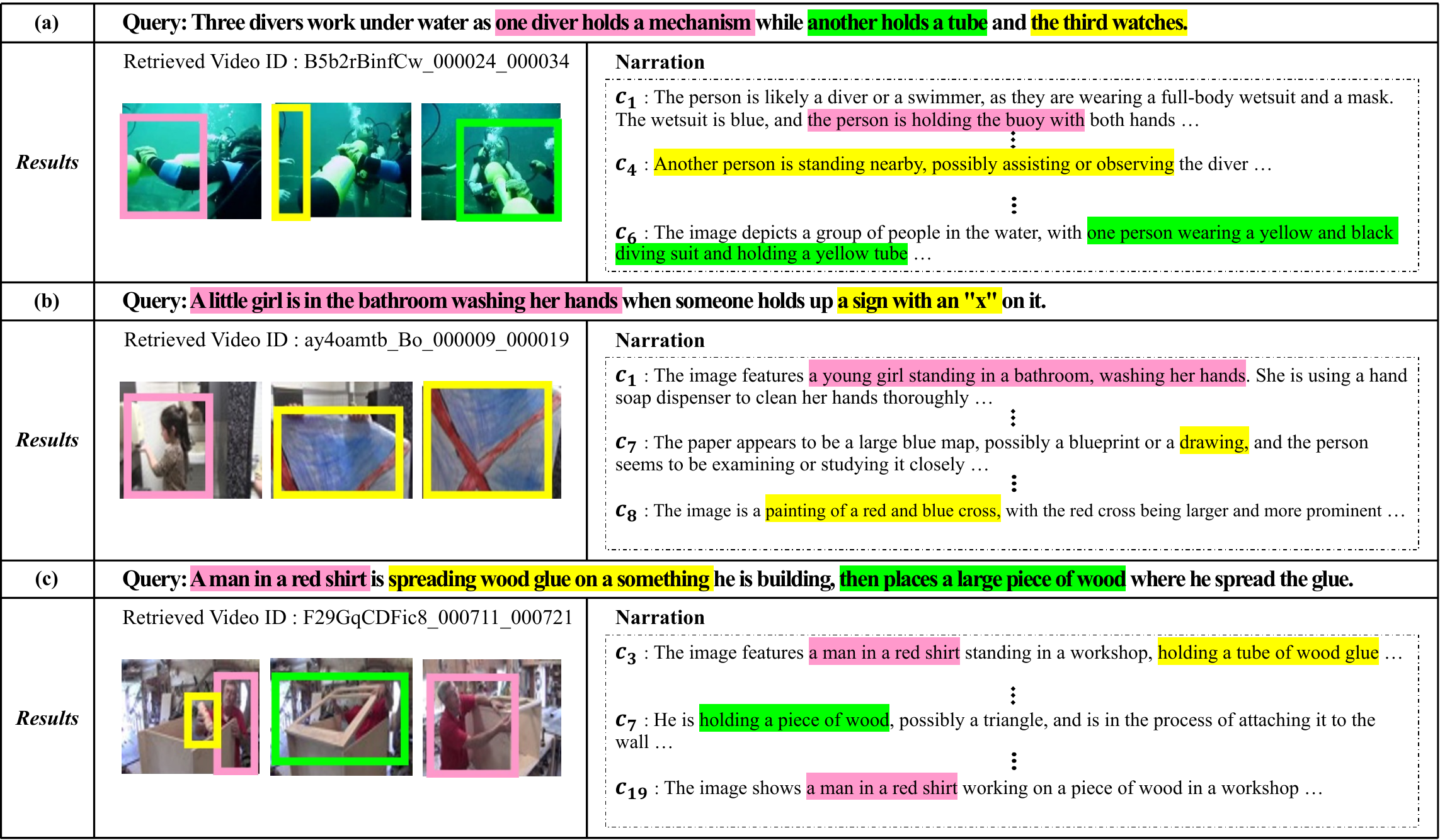}
    \caption{Text-to-video retrieval results on VATEX. (a) The narration effectively captures the actions of the three individuals in the video, leading to positive retrieval results for Narvid. (b) The narration effectively identifies local information in two scenes: One showing the girl washing her hands and another with her holding up the “x” sign. (c) Narvid correctly identifies the man in a red shirt by observing woodworking behaviors like using wood glue and handling wood pieces over time.}
    \label{fig:vatex_visualization}
\end{figure*}

\begin{figure*}
    \centering
    \includegraphics[width=0.90\linewidth]{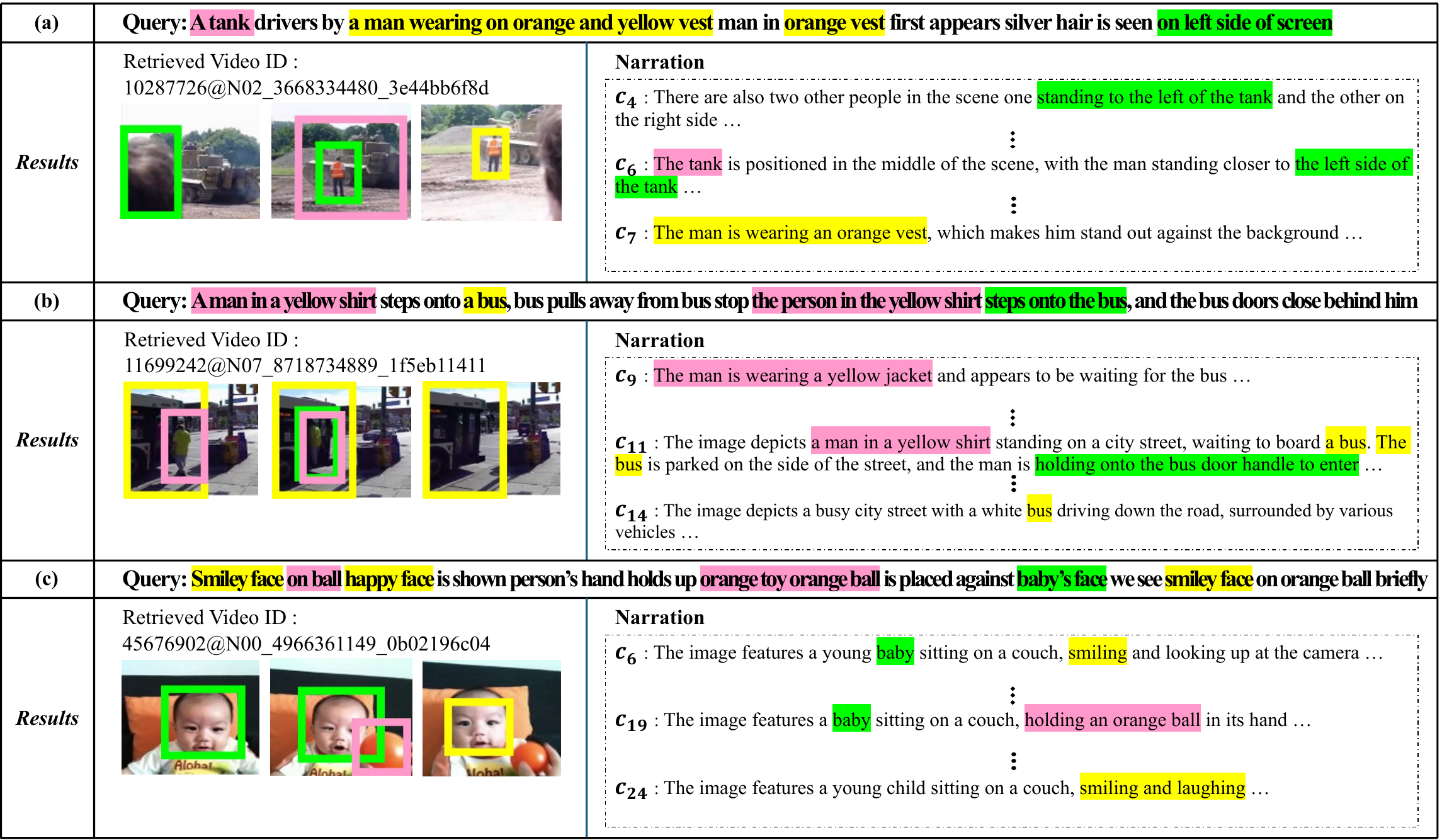}
    \caption{Text-to-video retrieval results on DiDeMo. (a) The narration is effective in identifying objects like a tank or a man in an orange vest in videos, as well as indicating location expressions like left. (b) NarVid clearly understands the behavior of the man wearing a yellow shirt over time and notes that he boards the bus. (c) The narration assists in identifying the baby, the ball, and the baby's joyful expression in the video.}
    \label{fig:didemo_visualization}
\end{figure*}

\begin{figure*}
    \centering
    \includegraphics[width=\linewidth]{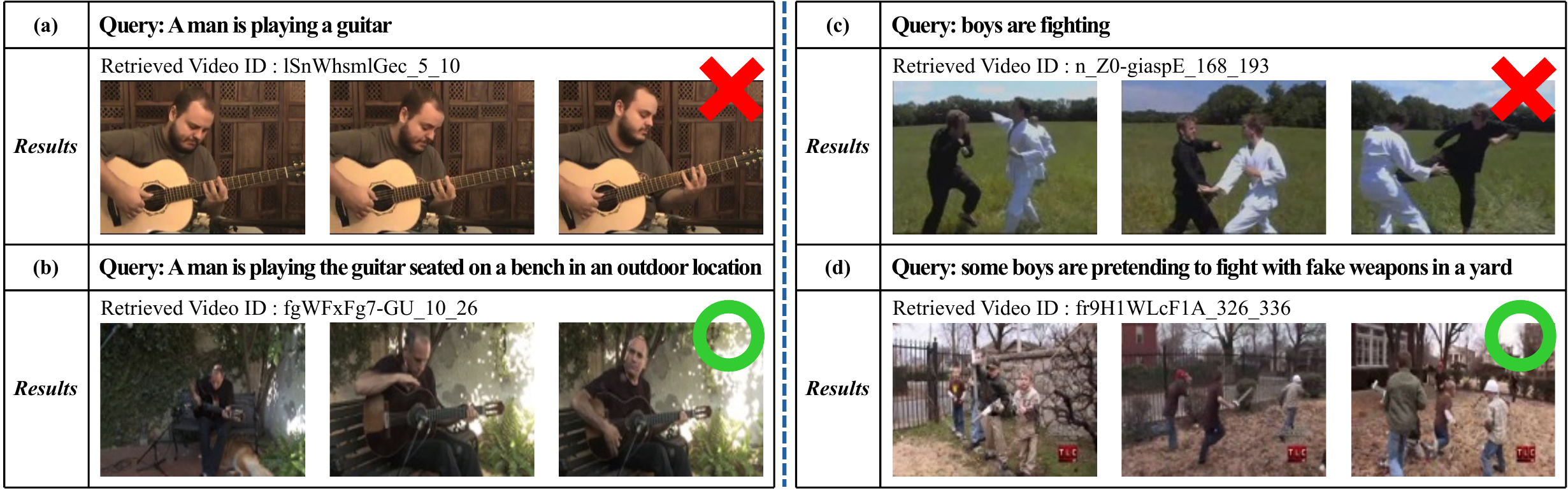}
    \caption{Example of mismatched results from NarVid due to a short query in the MSVD dataset: Queries (a) and (b) demonstrate how a short query with less information can lead to similar but incorrect retrieval results. In contrast, a detailed query specifying the location (sitting on a bench and outdoors) yields the correct answer. Similarly, queries (c) and (d) illustrate that adding specific details, such as holding weapons or fighting in the yard, to the short query (boys are fighting) can produce an accurate result.}
    \label{fig:msvd_short_query}
\end{figure*}

To qualitatively analyze and demonstrate the effectiveness of employing narration in text-video retrieval, we provide additional examples of retrieval results and generated frame-level captions for three datasets: \cref{fig:msvd_visualization} for MSVD, \cref{fig:vatex_visualization} for VATEX, and \cref{fig:didemo_visualization} for DiDeMo. Additionally, \cref{fig:msvd_short_query} shows examples with incorrect results due to short and general queries.


\newpage
% \nocite{*}
% {
%     \small
%     \bibliographystyle{ieeenat_fullname}
%     \bibliography{main}
% }

\end{document}